\documentclass[10pt,twocolumn,letterpaper]{article}

\usepackage{cvpr}
\usepackage{times}
\usepackage{epsfig}
\usepackage{graphicx}
\usepackage{amsmath}
\usepackage{amssymb}

\usepackage{mycommands}
\usepackage{color}
\usepackage{siunitx}
\usepackage{tabularx}
\usepackage{xpatch}
\usepackage{afterpage}

\usepackage[pagebackref=true,breaklinks=true,letterpaper=true,colorlinks,bookmarks=false]{hyperref}

\cvprfinalcopy %

 \newcommand{\GLD}{Google Landmarks Dataset v2}

\newcommand{\PAR}[1]{\vskip4pt \noindent {\bf #1}}

\def\cvprPaperID{2716} %

\begin{document}

\makeatletter
\DeclareRobustCommand\onedot{\futurelet\@let@token\@onedot}
\def\@onedot{\ifx\@let@token.\else.\null\fi\xspace}
\def\eg{\emph{e.g}\onedot} \def\Eg{\emph{E.g}\onedot}
\def\ie{\emph{i.e}\onedot} \def\Ie{\emph{I.e}\onedot}
\def\cf{\emph{cf}\onedot} \def\Cf{\emph{Cf}\onedot}
\def\etc{\emph{etc}\onedot} \def\vs{\emph{vs}\onedot}
\def\wrt{w.r.t\onedot} \def\dof{d.o.f\onedot}
\def\etal{\emph{et al}\onedot}
\def\vs{vs\onedot}
\makeatother

\interfootnotelinepenalty=10000

\title{Google Landmarks Dataset v2\\ \textit{A Large-Scale Benchmark for Instance-Level Recognition and Retrieval}}

\author{Tobias Weyand \thanks{equal contribution} \qquad Andr\'{e} Araujo \footnotemark[1] \qquad Bingyi Cao \qquad Jack Sim\\
Google Research, USA\\
{\tt\small \{weyand,andrearaujo,bingyi,jacksim\}@google.com}
}

\maketitle

\begin{abstract}
    While image retrieval and instance recognition techniques are progressing rapidly, there is a need for challenging datasets to accurately measure their performance -- while posing novel challenges that are relevant for practical applications.
    We introduce the \GLD{} (GLDv2), a new benchmark for large-scale, fine-grained instance recognition and image retrieval in the domain of human-made and natural landmarks. GLDv2 is the largest such dataset to date by a large margin, including over \num{5}M images and \num{200}k distinct instance labels. Its test set consists of \num{118}k images with ground truth annotations for both the retrieval and recognition tasks. The ground truth construction involved over 800 hours of human annotator work.
    Our new dataset has several challenging properties inspired by real-world applications that previous datasets did not consider: An extremely long-tailed class distribution, a large fraction of out-of-domain test photos and large intra-class variability.
    The dataset is sourced from Wikimedia Commons, the world's largest crowdsourced collection of landmark photos.
    We provide baseline results for both recognition and retrieval tasks based on state-of-the-art methods as well as competitive results from a public challenge. We further demonstrate the suitability of the dataset for transfer learning by showing that image embeddings trained on it achieve competitive retrieval performance on independent datasets.
    The dataset images, ground-truth and metric scoring code are available at \href{https://github.com/cvdfoundation/google-landmark}{https://github.com/cvdfoundation/google-landmark}.
\end{abstract}

\vspace{-10pt}

\section{Introduction}

Image retrieval and instance recognition are fundamental research topics which have been studied for decades. The task of image retrieval \cite{Philbin07,jegou2012aggregating,gordo2016deep,radenovic2018revisiting} is to rank images in an index set \wrt their relevance to a query image. The task of instance recognition \cite{kalantidis2011scalable,DelChiaro2019noisyart,liu2016deepfashion} is to identify which specific \emph{instance} of an object class (\eg the instance ``Mona Lisa" of the object class ``painting") is shown in a query image.

As techniques for both tasks have evolved, approaches have become more robust and scalable and are starting to ``solve" early datasets. Moreover, while increasingly large-scale classification datasets like ImageNet \cite{russakovsky2015imagenet}, COCO \cite{lin2014coco} and OpenImages \cite{kuznetsova2018open} have established themselves as standard benchmarks, image retrieval is still commonly evaluated on very small datasets. For example, the original Oxford\num{5}k \cite{Philbin07} and Paris\num{6}k \cite{Philbin2008} datasets that were released in \num{2007} and \num{2008}, respectively, have only \num{55} query images of \num{11} instances each, but are still widely used today. Because both datasets only contain images from a single city, results may not generalize to larger-scale settings.

\begin{figure}
    \centering
    \includegraphics[width=1\linewidth]{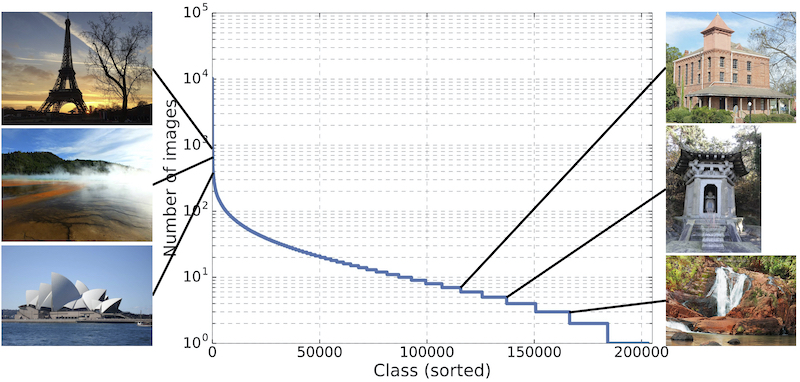}
    \vspace{-20pt}
    \caption[Caption for LOF]{The \GLD{} contains a variety of natural and human-made landmarks from around the world. Since the class distribution is very long-tailed, the dataset contains a large number of lesser-known local landmarks.\protect\footnotemark}
    \label{fig:example_images}
\end{figure}

\footnotetext{
    Photo attributions, top to bottom, left to right:
    \href{http://commons.wikimedia.org/wiki/File:Paris_le_matin_-_panoramio.jpg}{1} by fyepo, \href{http://creativecommons.org/licenses/by/3.0}{CC-BY},
    \href{http://commons.wikimedia.org/wiki/File:Yellowstone_Grand_Prismatic_Sunshine.jpg}{2} by C24winagain, \href{https://creativecommons.org/licenses/by-sa/4.0}{CC-BY-SA},
    \href{http://commons.wikimedia.org/wiki/File:The_Opera_House_-_panoramio.jpg}{3} by AwOiSoAk KaOsIoWa, \href{https://creativecommons.org/licenses/by-sa/4.0}{CC-BY-SA},
    \href{http://commons.wikimedia.org/wiki/File:Berrien_County_Jail,_Nashville,_GA,_US.jpg}{4} by Jud McCranie, \href{http://creativecommons.org/licenses/by-sa/4.0}{CC-BY-SA};
    \href{http://commons.wikimedia.org/wiki/File:Xin-yin-chan-shi-ta001.JPG}{5} by Shi.fachuang, \href{http://creativecommons.org/licenses/by-sa/1.0}{CC-BY-SA};
    \href{http://commons.wikimedia.org/wiki/File:Hang_Cop_Waterfall_in_Da_Lat_1.jpg}{6} by Nhi Dang, \href{http://creativecommons.org/licenses/by/2.0}{CC-BY}.}

Many existing datasets also do not present real-world challenges. For instance, a landmark recognition system that is applied in a generic visual search app will be queried with a large fraction of non-landmark queries, like animals, plants, or products, which it is not expected to yield any results for. Yet, most instance recognition datasets have only “on-topic” queries and do not measure the false-positive rate on out-of-domain queries. Therefore, larger, more challenging datasets are necessary to fairly benchmark these techniques while providing enough challenges to motivate further research.

\begin{figure}
    \centering
    \includegraphics[width=1\linewidth]{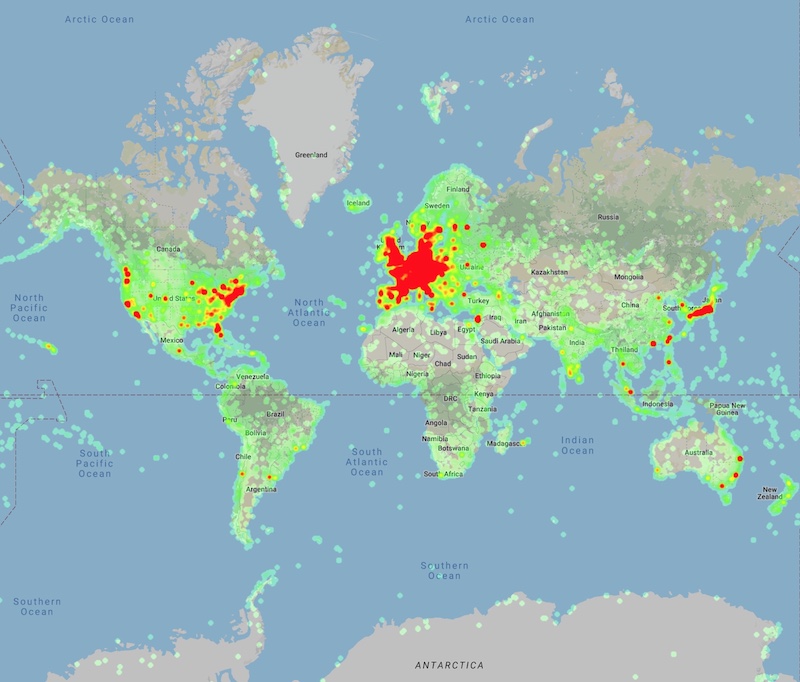}
    \vspace{-15pt}
    \caption{Heatmap of the places in the \GLD{}.}
    \vspace{-5pt}
    \label{fig:map}
\end{figure}

A possible reason that small-scale datasets have been the dominant benchmarks for a long time is that it is hard to collect instance-level labels at scale. Annotating millions of images with hundreds of thousands of fine-grained instance labels is not easy to achieve when using labeling services like Amazon Mechanical Turk, since taggers need expert knowledge of a very fine-grained domain.

We introduce the \GLD{} (GLDv2), a new large-scale dataset for instance-level recognition and retrieval. GLDv2 includes over \num{5}M images of over \num{200}k human-made and natural landmarks that were contributed to Wikimedia Commons by local experts. \figref{fig:example_images} shows a selection of images from the dataset and \figref{fig:map} shows its geographical distribution. The dataset includes \num{4}M labeled training images for the instance recognition task and \num{762}k index images for the image retrieval task. The test set consists of \num{118}k query images with ground truth labels for both tasks. To mimic a realistic setting, only \num{1}\% of the test images are within the target domain of landmarks, while \num{99}\% are out-of-domain images. While the \GLD{} focuses on the task of recognizing landmarks, approaches that solve the challenges it poses should readily transfer to other instance-level recognition tasks, like logo, product or artwork recognition.

The \GLD{} is designed to simulate real-world conditions and thus poses several hard challenges. It is \emph{large scale} with millions of images of hundreds of thousands of classes. The distribution of these classes is very long-tailed (\figref{fig:example_images}), making it necessary to deal with extreme \emph{class imbalance}. The test set has a large fraction of \emph{out-of-domain images}, emphasizing the need for low false-positive recognition rates. The \emph{intra-class variability} is very high, since images of the same class can include indoor and outdoor views, as well as images of indirect relevance to a class, such as paintings in a museum. The goal of the \GLD{} is to become a new benchmark for instance-level recognition and retrieval. In addition, the recognition labels can be used for training image descriptors or pre-training approaches for related domains where less data is available. We show that the dataset is suitable for transfer learning by applying learned descriptors on independent datasets where they achieve competitive performance.

The dataset was used in two public challenges on Kaggle\footnote{\href{https://www.kaggle.com/c/landmark-recognition-2019}{https://www.kaggle.com/c/landmark-recognition-2019}, \href{https://www.kaggle.com/c/landmark-retrieval-2019}{https://www.kaggle.com/c/landmark-retrieval-2019}}, where researchers and hobbyists competed to develop models for instance recognition and image retrieval. We discuss the results of the challenges in \secref{sec:experiments}.

The dataset images, instance labels for training, the ground truth for retrieval and recognition and the metric computation code are publicly available\footnote{\href{https://github.com/cvdfoundation/google-landmark}{https://github.com/cvdfoundation/google-landmark}}.

\section{Related Work}

\newcolumntype{Y}{>{\centering\arraybackslash}X}
\begin{table*}
\footnotesize
\addtolength{\tabcolsep}{-0.15em}
\centering
{%
\begin{tabularx}{\textwidth}{l c c Y Y Y c c c}
Dataset name  & Year & \# Landmarks & \# Test \newline images & \# Train \newline images & \# Index images & Annotation collection & Coverage & Stable \\
\midrule
Oxford \cite{Philbin07} & 2007 & 11 & 55 & - & 5k & Manual & City & Y \\
Paris \cite{Philbin2008} & 2008 & 11 & 55 & - & 6k & Manual & City & Y \\
Holidays \cite{Jegou2008} & 2008 & 500 & 500 & - & 1.5k & Manual & Worldwide & Y \\
European Cities \num{50}k \cite{avrithis2010feature} & 2010 & 20 & 100 & - & 50k & Manual & Continent & Y \\
Geotagged StreetView \cite{knopp2010avoiding} & 2010 & - & 200 & - & 17k & StreetView & City & Y \\
Rome 16k \cite{agarwal2011building} & 2010 & 69 & 1k & - & 15k & GeoTag + SfM & City & Y \\
San Francisco \cite{chen2011city} & 2011 & - & 80 & - & \bf{1.7M} & StreetView & City & Y \\
Landmarks-PointCloud \cite{li2012worldwide} & 2012 & 1k & 10k & - & 205k & Flickr label + SfM & Worldwide & Y \\
24/7 Tokyo \cite{torii201524} & 2015 & 125 & 315 & - & 1k & Smartphone + Manual & City & Y \\
Paris500k \cite{Weyand2015} & 2015 & 13k & 3k & - & 501k & Manual & City & N \\
Landmark URLs \cite{babenko2014neural,gordo2016deep} & 2016 & 586 & - & 140k & - & Text query + Feature matching & Worldwide & N \\
Flickr-SfM \cite{radenovic2018fine} & 2016 & 713 & - & 120k & - & Text query + SfM & Worldwide & Y \\
Google Landmarks \cite{noh2017large} & 2017 & 30k & \bf{118k} & 1.2M & 1.1M & GPS + semi-automatic & Worldwide & N \\
Revisited Oxford \cite{radenovic2018revisiting} & 2018 & 11 & 70 & - & 5k + 1M & Manual + semi-automatic & Worldwide & Y \\
Revisited Paris \cite{radenovic2018revisiting} & 2018 & 11 & 70 & - & 6k + 1M & Manual + semi-automatic & Worldwide & Y \\
\midrule
\GLD{} & 2019 & \bf{200k} & \bf{118k} & \bf{4.1M} & 762k & Crowsourced + semi-automatic & Worldwide & Y \\
\end{tabularx}
}
\vspace{-5pt}
\caption{Comparison of our dataset against existing landmark recognition/retrieval datasets. ``Stable'' denotes if the dataset can be retained indefinitely. Our \GLD{} is larger than all existing datasets in terms of total number of images and landmarks, besides being stable.}
\label{tab:landmark_datasets}
\end{table*}

Image recognition problems range from basic categorization (``cat'', ``shoe'', ``building''), through fine-grained tasks involving distinction of species/models/styles (``Persian cat'', ``running shoes'', ``Roman Catholic church''), to instance-level recognition (``Oscar the cat'', ``Adidas Duramo 9'', ``Notre-Dame cathedral in Paris'').
Our new dataset focuses on tasks that are at the end of this continuum: identifying individual human-made and natural landmarks.
In the following, we review image recognition and retrieval datasets, focussing mainly on those which are most related to our work.

\PAR{Landmark recognition/retrieval datasets.}
We compare existing datasets for landmark recognition and retrieval against our newly-proposed dataset in \tabref{tab:landmark_datasets}.
The Oxford \cite{Philbin07} and Paris \cite{Philbin2008} datasets contain tens of query images and thousands of index images from landmarks in Oxford and Paris, respectively.
They have consistently been used in image retrieval for more than a decade, and were re-annotated recently, with the addition of \num{1}M worldwide distractor index images \cite{radenovic2018revisiting}.
Other datasets also focus on imagery from a single city: Rome \num{16}k \cite{agarwal2011building}; Geotagged Streetview Images \cite{knopp2010avoiding} containing \num{17}k images from Paris; San Francisco Landmarks \cite{chen2011city} containing \num{1.7}M images; 24/7 Tokyo \cite{torii201524} containing \num{1}k images under different illumination conditions and Paris\num{500}k \cite{Weyand2015}, containing \num{501}k images.

More recent datasets contain images from a much larger variety of locations.
The European Cities (EC) \num{50}k dataset \cite{avrithis2010feature} contains images from \num{9} cities, with a total of \num{20} landmarks; unannotated images from other \num{5} cities are used as distractors.
This dataset also has a version with \num{1}M images from \num{22} cities where the annotated images come from a single city \cite{avrithis2010retrieving}.
The Landmarks dataset by Li \etal \cite{li2012worldwide} contains \num{205}k images of \num{1}k famous landmarks.
Two other recent landmark datasets, by Gordo \etal \cite{gordo2016deep} and Radenovic \etal \cite{radenovic2018fine}, have become popular for training image retrieval models, containing hundreds of landmarks and approximately \num{100}k images each; note that these do not contain test images, but only training data.
The original Google Landmarks Dataset \cite{noh2017large} contains \num{2.3}M images from \num{30}k landmarks, but due to copyright restrictions this dataset is not stable: it shrinks over time as images get deleted by the users who uploaded them.
The \GLD{} dataset surpasses all existing datasets in terms of the number of images and landmarks, and uses images only with licenses that allow free reproduction and indefinite retention.

\PAR{Instance-level recognition datasets.}
Instance-level recognition refers to a very fine-grained identification problem, where the goal is to visually recognize a single (or indistinguishable) occurrence of an entity.
This problem is typically characterized by a large number of classes, with high imbalance, and small intra-class variation.
Datasets for such problems have been introduced in the community, besides the landmark datasets mentioned previously.
For example: 
logos \cite{fehervari2019scalable,joly2009logo,kalantidis2011scalable,romberg2011scalable}, 
cars \cite{bai2018group,yan2017exploiting,zapletal2016vehicle}, 
products \cite{liu2016deepfashion,ge2019deepfashion2,wei2019rpc,song2016deep}, 
artwork \cite{arandjelovic2011smooth,DelChiaro2019noisyart}, 
among others \cite{chandrasekhar2011the}.

\PAR{Other image recognition datasets.}
There are numerous computer vision datasets for other types of image recognition problems.
Basic image categorization is addressed by datasets such as Caltech 101 \cite{feifei2004learning}, Caltech 256 \cite{griffin2007}, ImageNet \cite{russakovsky2015imagenet} and more recently OpenImages \cite{kuznetsova2018open}.
Popular fine-grained recognition datasets include CUB-200-2011 \cite{WahCUB_200_2011}, iNaturalist \cite{vanhorn2018the}, Stanford Cars \cite{krause20133d}, Places \cite{zhou2017places}.

\section{Dataset Overview}

\subsection{Goals}
The \GLD{} aims to mimic the following challenges of industrial landmark recognition systems: \emph{Large scale}: To cover the entire world, a corpus of millions of photos is necessary. \emph{Intra-class variability}: Photos are taken under varying lighting conditions and from different views, including indoor and outdoor views of buildings. There will also be photos related to the landmark, but not showing the landmark itself, \eg floor plans, portraits of architects, or views \emph{from} the landmark. \emph{Long-tailed class distribution}: There are much more photos of famous landmarks than of lesser-known ones. \emph{Out-of-domain queries}: The query stream that these systems receive may come from various applications such as photo album apps or visual search apps and often contains only a small fraction of landmarks among many photos of other object categories. This poses a significant challenge for the robustness of the recognition algorithm. We designed our dataset to capture these challenges. An additional goal was to use only images whose licenses permit indefinite retention and reproduction in publications.

\PAR{Non-goals.}
In contrast to many other datasets, we explicitly did not design GLDv2 to have clean query and index sets for the reasons mentioned above.
Also, the dataset does not aim to measure generalization of embedding models to unseen data -- therefore, the index and training sets do not have disjoint class sets.
Finally, we do not aim to provide an image-level retrieval ground truth at this point due to very expensive annotation costs. Instead, the retrieval ground truth is on a class-level, \ie all index images that belong to the same class as a query image will be marked as relevant in the ground truth.

\subsection{Scale and Splits}
The \GLD{} consists of over \num{5}M images and over \num{200}k distinct instance labels, making it the largest instance recognition dataset to date. It is divided into three subsets: (i) \num{118}k \emph{query} images with ground truth annotations, (ii) \num{4.1}M \emph{training} images of \num{203}k landmarks with labels that can be used for training, and (iii) \num{762}k \emph{index} images of \num{101}k landmarks. We also make available a cleaner, reduced training set of \num{1.6}M images and \num{81}k landmarks (see \secref{subsec:train_preprocessing}). While the index and training set do not share images, their label space is highly overlapping, with \num{92}k common classes. The query set is randomly split into 1/3 validation and 2/3 testing data. The validation data was used for the ``Public" leaderboard in the Kaggle competition, which allowed participants to submit solutions and view their scores in real-time. The test set was used for the ``Private" leaderboard, which was used for the final ranking and was only revealed at the end of the competition.

\subsection{Challenges}
Besides its scale, the \GLD{} presents practically relevant challenges, as motivated above.

\noindent \textbf{Class distribution.}
The class distribution is extremely long-tailed, as illustrated in \figref{fig:example_images}. \num{57}\% of classes have at most \num{10} images and \num{38}\% of classes have at most \num{5} images. The dataset therefore contains a wide variety of landmarks, from world-famous ones to lesser-known, local ones.

\noindent \textbf{Intra-class variation.}
As is typical for an image dataset collected from the web, the \GLD{} has large intra-class variability, including views from different vantage points and of different details of the landmarks, as well as both indoor and outdoor views for buildings.

\noindent \textbf{Out-of-domain query images.}
To simulate a realistic query stream, the query set consists of only \num{1.1}\% images of landmarks and \num{98.9}\% out-of-domain images, for which no result is expected. This puts a strong emphasis on the importance of robustness in a practical instance recognition system. %

\subsection{Metrics}
The \GLD{} uses well-established metrics, which we now introduce. Reference implementations are available on the dataset website.

\PAR{Recognition}
is evaluated using micro Average Precision ($\mu$AP) \cite{Perronnin09CVPR} with one prediction per query. This is also known as Global Average Precision (GAP). It is calculated by sorting all predictions in descending order of their confidence and computing:
\vspace{-7px}
\begin{equation}
\mu\mathrm{AP} = \frac{1}{M} \sum_{i=1}^N \mathrm{P}(i) \mathrm{rel}(i) ,
\vspace{-7px}
\end{equation}
where $N$ is the total number of predictions across all queries; $M$ is the total number of queries with at least one landmark from the training set visible in it (note that most queries do not depict landmarks); $\mathrm{P}(i)$ is the precision at rank $i$; and $\mathrm{rel}(i)$ is a binary indicator function denoting the correctness of prediction $i$.
Note that this metric penalizes a system that predicts a landmark for an out-of-domain query image; overall, it measures both ranking performance as well as the ability to set a common threshold across different queries.

\PAR{Retrieval}
is evaluated using mean Average Precision@100 (mAP@100), which is a variant of the standard mAP metric that only considers the top-100 ranked images. We chose this limitation since exhaustive retrieval of every matching image is not necessary in most applications, like image search. The metric is computed as follows:
\vspace{-5px}
\begin{equation}
\vspace{-10px}
\mathrm{mAP@100} =  \frac{1}{Q} \sum_{q=1}^{Q} \mathrm{AP@100(q)} ,
\end{equation}
where
\begin{equation}
\mathrm{AP@100(q)} = \frac{1}{\mathrm{min}(m_q, 100)} \sum_{k=1}^{\mathrm{min}(n_q, 100)} \mathrm{P}_q(k) \mathrm{rel}_q(k)
\end{equation}

\noindent where $Q$ is the number of query images that depict landmarks from the index set; $m_q$ is the number of index images containing a landmark in common with the query image $q$ (note that this is only for queries which depict landmarks from the index set, so $m_q\neq0$); $n_q$ is the number of predictions made by the system for query $q$; $\mathrm{P}_q(k)$ is the precision at rank $k$ for the $q$-th query; and $\mathrm{rel}_q(k)$ is a binary indicator function denoting the relevance of prediction $k$ for the $q$-th query. Some query images will have no associated index images to retrieve; these queries are ignored in scoring, meaning this metric does not penalize the system if it retrieves landmark images for out-of-domain queries.

\begin{figure}
    \centering
    \includegraphics[width=1.0\linewidth]{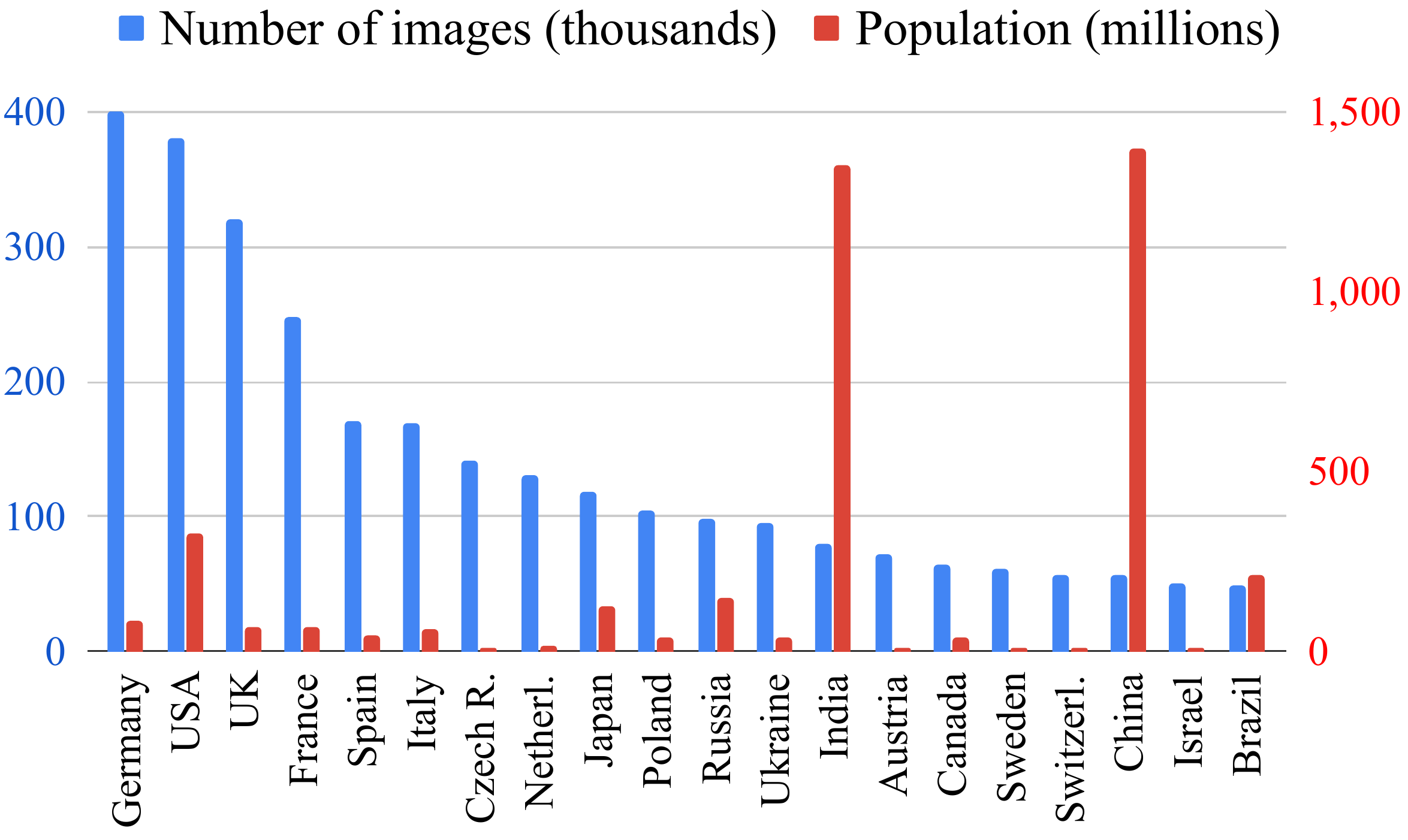}
    \vspace{-20pt}
    \caption{Histogram of the number of images from the top-20 countries (blue) compared to their populations (red).}
    \label{fig:country_distribution}
\end{figure}

\subsection{Data Distribution}
\begin{figure}
    \centering
    \includegraphics[width=1.0\linewidth]{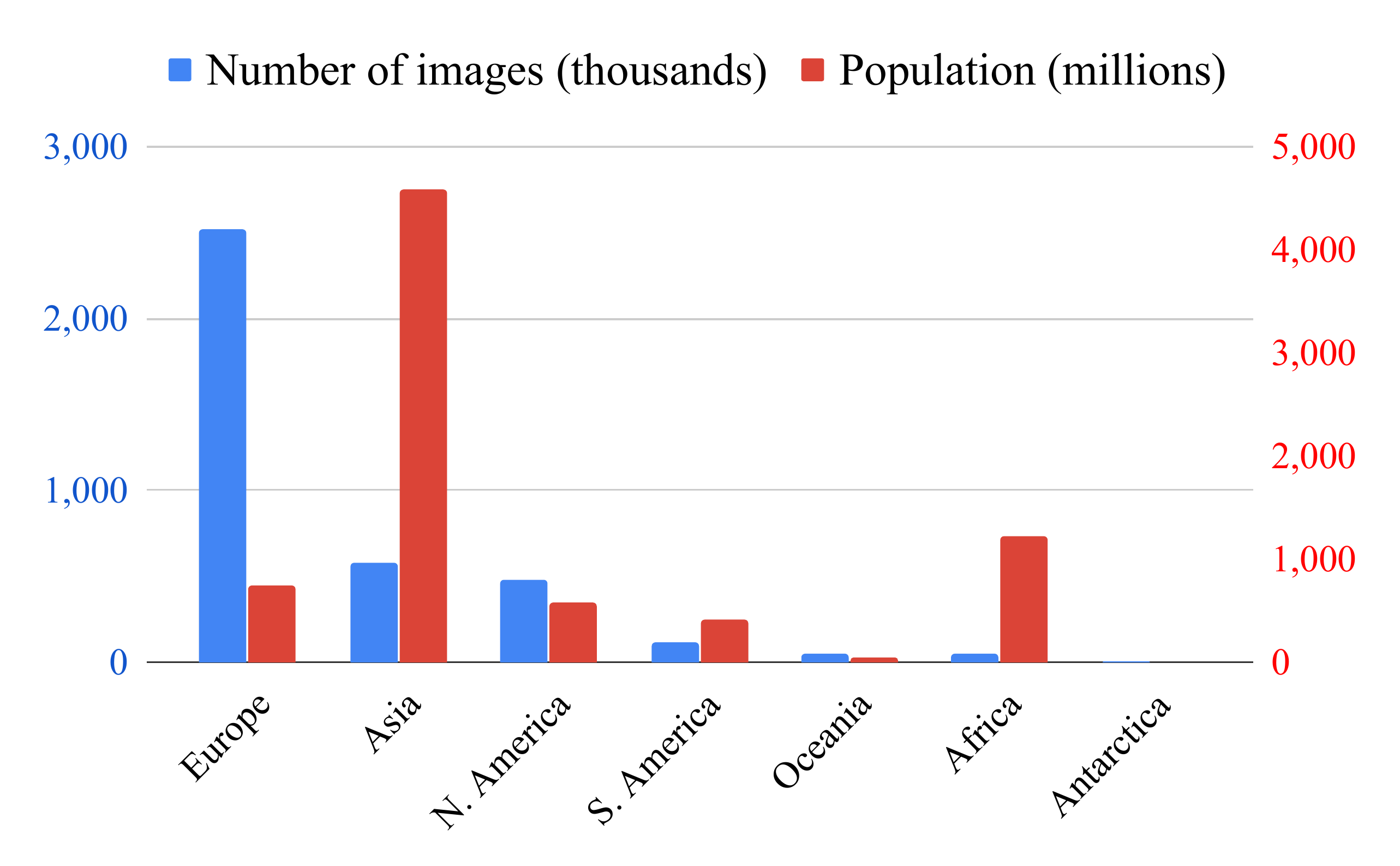}
        \vspace{-25pt}
    \caption{Histogram of the number of images per continent (blue) compared to their populations (red).}
    \label{fig:continent_distribution}
\end{figure}

The \GLD{} is a truly world-spanning dataset, containing landmarks from 246 of the 249 countries in the ISO 3166-1 country code list. \figref{fig:country_distribution} shows the number of images in the top-20 countries and \figref{fig:continent_distribution} shows the number of images by continent. We can see that even though the dataset is world-spanning, it is by no means a representative sample of the world, because the number of images per country depends heavily on the activity of the local Wikimedia Commons community.

\begin{figure}
    \centering
    \includegraphics[width=1.0\linewidth]{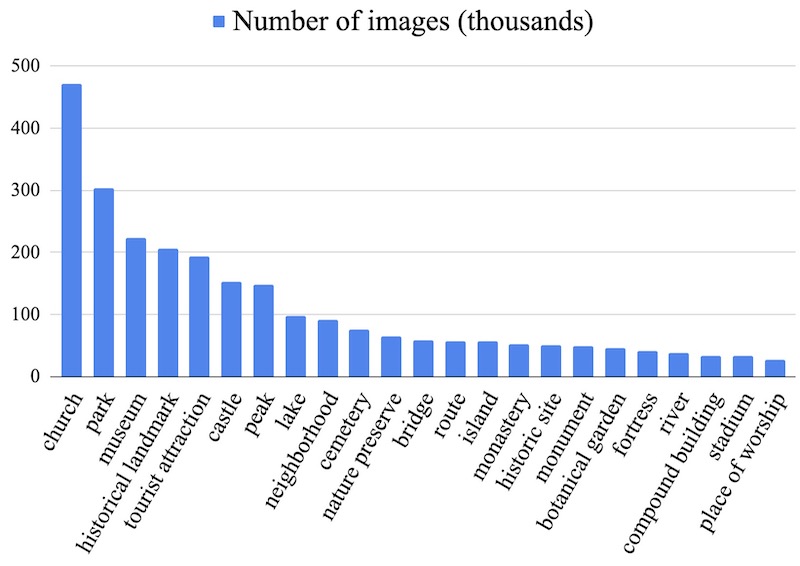}
    \vspace{-20pt}
    \caption{Histogram of the number of images by landmark category. This includes only categories with more than \num{25}k images.}
    \label{fig:category_distribution}
\end{figure}

\figref{fig:category_distribution} shows the distribution of the dataset images by landmark category, as obtained from the Google Knowledge Graph. By far the most frequent category is churches, followed by parks and museums. Counting only those categories with over \num{25}k images, roughly \num{28}\% are natural landmarks while \num{72}\% are human-made.

\subsection{Image Licenses}

All images in GLDv2 are freely licensed, so that the dataset  is indefinitely retainable and does not shrink over time, allowing recognition and retrieval approaches to be compared over a long period of time. All images can be freely reproduced in publications, as long as proper attribution is provided. The image licenses are either Creative Commons\footnote{\href{https://creativecommons.org}{https://creativecommons.org}} or Public Domain. We provide a list of attributions for those images that require it so dataset users can easily give attribution when using the images in print or on the web.

\begin{figure*}[t]
\centering
\begin{subfigure}{0.65\textwidth}
  \centering
    \includegraphics[width=0.95\linewidth]{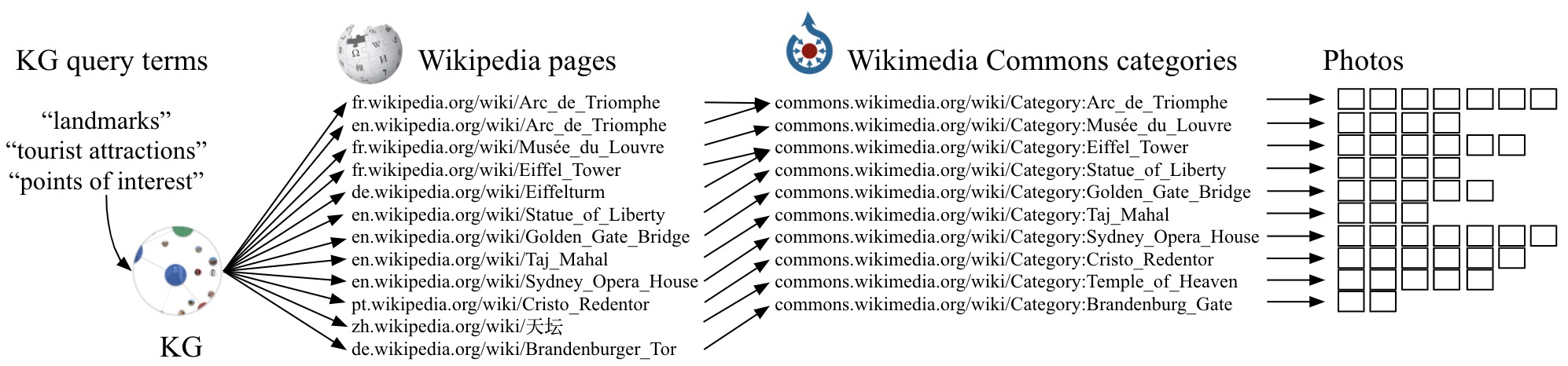}
\end{subfigure}%
\begin{subfigure}{0.35\textwidth}
    \includegraphics[width=1\linewidth]{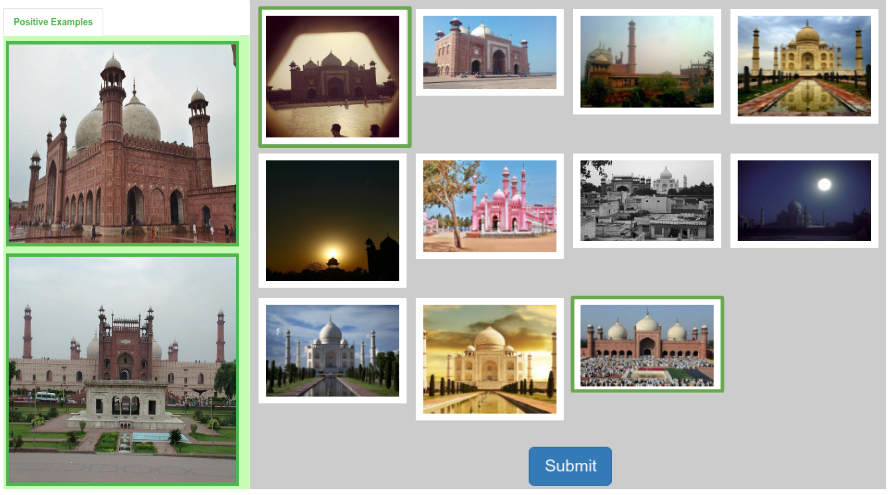}    
\end{subfigure}%
\vspace{-8pt}
\caption{Left: pipeline for mining images from Wikimedia Commons. Right: the user interface of the re-annotation tool.}
\label{fig:wmc_mining_and_reannotation_ui}
\end{figure*}

\section{Dataset Construction}
This section details the data collection process and the construction of the ground truth.

\subsection{Data Collection}
\paragraph{Data sources.}
The main data source of the \GLD{} is \emph{Wikimedia Commons}\footnote{\href{https://commons.wikimedia.org}{https://commons.wikimedia.org}}, the media repository behind Wikipedia. Wikimedia Commons hosts millions of photos of landmarks licensed under Creative Commons and Public Domain licenses, contributed by an active community of photographers as well as partner organizations such as libraries, archives and museums. Its large coverage of the world's landmarks is partly thanks to \emph{Wiki Loves Monuments}\footnote{\href{https://www.wikilovesmonuments.org}{https://www.wikilovesmonuments.org}}, an annual world-wide contest with the goal to upload high-quality, freely licensed photos of landmarks to the site and to label them within a fine-grained taxonomy of the world's cultural heritage sites.
In addition to Wikimedia Commons, we collected realistic query images by crowdsourcing. Operators were sent out to take photos of selected landmarks around the world with smartphones.

\vspace{-7pt}

\paragraph{Training and index sets.}
\figref{fig:wmc_mining_and_reannotation_ui} (left) shows the process we used to mine landmark images from Wikimedia Commons. Wikimedia Commons is organized into a hierarchy of categories that form its taxonomy. Each category has a unique URL where all its associated images are listed. We found that the Wikimedia Commons hierarchy does not have a suitable set of top-level categories that map to human-made and natural landmarks. Instead, we found the Google Knowledge Graph\footnote{\href{https://developers.google.com/knowledge-graph}{https://developers.google.com/knowledge-graph}} to be useful to obtain an exhaustive list of the landmarks of the world. To obtain a list of Wikimedia Commons categories for landmarks, we queried the Google Knowledge Graph with terms like ``landmarks", ``tourist attractions", ``points of interest", \etc. For each returned knowledge graph entity, we obtained its associated Wikipedia articles. We then followed the link to the Wikimedia Commons Category page in the Wikipedia article. Note that while Wikipedia may have articles about the same landmark in different languages, Wikimedia Commons only has one category per subject. We then downloaded all images contained in the Wikimedia Commons pages we obtained. To avoid ambiguities, we enforced the restriction that each mined image be associated to a single Wikimedia category or Knowledge Graph entity. We use the Wikimedia Commons category URLs as the canonical class labels. The training and index sets are collected in this manner.

\vspace{-10pt}

\paragraph{Query set.}
The query set consists of ``positive" query images of landmarks and ``negative" query images not showing landmarks.
For collecting the ``positive" query set, we selected a subset of the landmarks we collected from Wikimedia Commons and asked crowdsourcing operators to take photos of them.
For the ``negative" query data collection, we used the same process as for the index and training data, but queried the Knowledge Graph only with terms that are unrelated to landmarks. We also removed any negative query images that had near-duplicates in the index or training sets.

\vspace{-10pt}

\paragraph{Dataset partitioning.}

The landmark images from Wikimedia Commons were split into training and index sets based on their licenses. We used CC0 and Public Domain photos for the index set while photos with ``Creative Commons By" licenses that did not have a ``No Derivatives" clause were used for the training set. As a result, the label spaces of index and training sets have a large, but not complete, overlap.

\subsection{Test Set Re-Annotation}
\label{sec:reannotation}
Visual inspection of retrieval and recognition results showed that many errors were due to missing ground truth annotations, which was due to the following reasons: (i) Crowdsourced labels from Wikimedia Commons can contain errors and omissions. (ii) Some query images contain multiple landmarks, but only one of them was present in the ground truth. (iii) There are sometimes multiple valid labels for an image on different hierarchical levels. For example, for a picture of a mountain in a park, both the mountain and the park would be appropriate labels. (iv) Some of the ``negative" query images do actually depict landmarks.

We therefore amend the ground truth with human annotations. However, the large number of instance labels makes this a challenging problem: Each query image would need to be annotated with one out of \num{200}k landmark classes, which is infeasible for human raters. We therefore used the model predictions of the top-ranked teams from the challenges to propose potential labels to the raters. To avoid bias in the new annotation towards any particular method, we used the top-\num{10} submissions which represent a wide range of methods (see \secref{sec:challenge_results}). A similar idea was used to construct the distractor set of the revisited Oxford and Paris datasets \cite{radenovic2018revisiting}, where hard distractors were mined using a combination of different retrieval methods.

\figref{fig:wmc_mining_and_reannotation_ui} (right) shows the user interface of the re-annotation tool.
On the left side, we show a sample of index/train images of a given landmark label. On the right, we show the query images that are proposed for this label and ask raters to click on the correct images. This way, we simplified the question of ``which landmark is it?" as ``is it this landmark?", which is a simple ``yes" or ``no" question. Grouping the query images associated with the same landmark class together further improved the re-annotation efficiency, since raters do not need to switch context between landmark classes. To make efficient use of rater time, we only selected the highest-confidence candidates from the top submissions, since those are more likely to be missing annotations rather than model errors. In total, we sent out $\sim$\num{10}k recognition query images and $\sim$\num{90}k retrieval query images for re-annotation. To ensure the re-annotation quality, we sent each image to \num{3} human raters and assigned the label based on majority voting. In total, we leveraged $\sim$\num{800} rater hours on the re-annotation process. This re-annotation increased the number of recognition annotations by \num{72}\% and the number of retrieval annotations by \num{30}\%. If a ``negative" query image was verified to contain a landmark, it was moved to the ``positive" query set. We will continue to improve the ground truth and will make future versions available on the dataset website. For comparability, past versions will stay available and each ground truth will receive a version number that we ask dataset users to state when publishing results.

\section{Experiments}
\label{sec:experiments}

We demonstrate usage of the dataset and present several baselines that can be used as reference results for future research, besides discussing results from the public challenge. All results presented in this section are \wrt version 2.1 of the dataset ground truth.

\subsection{Training Set Pre-Processing} \label{subsec:train_preprocessing}

The \GLD{} training set presents a realistic crowdsourced setting with diverse types of images for each landmark: \eg, for a specific museum there may be outdoor images showing the building facade, but also indoor images of paintings and sculptures that are on display.
Such diversity within a  class may pose challenges to the training process, so we consider the pre-processing steps proposed in \cite{Yokoo20Landmark} in order to make each class more visually coherent.
Within each class, each image is queried against all others by global descriptor similarity, followed by geometric verification of the top-$100$ most similar images using local features.
The global descriptor is a ResNet-101 \cite{he2015deep} embedding and the local features are DELF \cite{noh2017large}, both trained on the first Google Landmarks Dataset version (GLDv1) \cite{noh2017large}.
If an image is successfully matched to at least $3$ other images, each with at least $30$ inliers, it is selected; otherwise discarded.

We refer to the resulting dataset version as GLDv2-train-clean and make it available on the dataset website.
\tabref{tab:cleaned_train_set} presents the number of selected images and labels: $1.6$M training images ($38$\%) and $81$k labels ($40$\%).
Even if this version only contains less than half of the data from GLDv2-train, it is still much larger than the training set of any other landmark recognition dataset.
We also experiment with a variant of GLDv2-train-clean, where classes with fewer than $15$ images are removed, referred to as GLDv2-train-no-tail; it has approximately the same number of images as GLDv1-train, but $2\times$ the number of classes.

\begin{table}
\footnotesize
\addtolength{\tabcolsep}{-0.15em}
\centering
{%
\begin{tabular}{l c c}
Training set  & \# Images & \# Labels \\
\midrule
GLDv1-train \cite{noh2017large} & $1,225,029$ & $14,951$ \\
GLDv2-train & $4,132,914$ & $203,094$ \\
GLDv2-train-clean & $1,580,470$ & $81,313$ \\
GLDv2-train-no-tail & $1,223,195$ & $27,756$ \\
\end{tabular}
}
\vspace{-5pt}
\caption{Number of images and labels for the different training datasets used in our experiments.}
\label{tab:cleaned_train_set}
\end{table}

\begin{table}[t]
\setlength{\tabcolsep}{1pt}
\footnotesize
\centering
\begin{tabular}{l c c c c c}
 \multirow{2}{*}{Technique} &  \multirow{2}{*}{Training Dataset} & \multicolumn{2}{c}{Medium} & \multicolumn{2}{c}{Hard} \\ 
 & & \multicolumn{1}{c}{$\mathcal{R}$Oxf} & \multicolumn{1}{c}{$\mathcal{R}$Par} & \multicolumn{1}{c}{$\mathcal{R}$Oxf} & \multicolumn{1}{c}{$\mathcal{R}$Par} \\
\midrule
\multirow{5}{*}{ResNet101+ArcFace} & Landmarks-full \cite{gordo2016deep} & $50.8$ & $70.4$ & $24.3$ & $47.1$ \\
 & Landmarks-clean \cite{gordo2016deep} & $54.2$ & $70.7$ & $28.3$ & $46.0$ \\
 & GLDv1-train \cite{noh2017large} & $68.9$ & $83.4$ & $45.3$ & $67.2$ \\
 & GLDv2-train-clean & $76.2$ & $\mathbf{87.3}$ & $55.6$ & $\textbf{74.2}$ \\ 
 & GLDv2-train-no-tail & $76.1$ & $86.6$ & $55.1$ & $72.5$ \\
\hline
DELF-ASMK*+SP \cite{radenovic2018revisiting} & \multirow{7}{*}{GLDv1-train \cite{noh2017large}} & $67.8$ & $76.9$ & $43.1$ & $55.4$ \\
DELF-R-ASMK* \cite{teichmann2019d2r} & &$73.3$ & $80.7$ & $47.6$ & $61.3$ \\
DELF-R-ASMK*+SP \cite{teichmann2019d2r} && $76.0$ & $80.2$ & $52.4$ & $58.6$ \\
ResNet152+Triplet \cite{radenovic2018fine} & & $68.7$ & $79.7$ & $44.2$ & $60.3$ \\
ResNet101+AP \cite{revaud2019aploss} & & $66.3$ & $80.2$ & $42.5$ & $60.8$ \\
DELG global-only \cite{Cao2020DELG} && $73.2$ & $82.4$ & $51.2$ & $64.7$ \\
DELG global+SP \cite{Cao2020DELG} & & $\mathbf{78.5}$ & $82.9$ & $\mathbf{59.3}$ & $65.5$ \\
\hline
\multicolumn{2}{l}{HesAff-rSIFT-ASMK*+SP \cite{radenovic2018revisiting} \hspace{20pt}-} & $60.6$ & $61.4$ & $36.7$ & $35.0$ \\
\end{tabular}
\vspace{-5pt}
\caption{Retrieval results (\% mAP) on $\mathcal{R}$Oxf and $\mathcal{R}$Par of baseline models (ResNet-101 + ArcFace loss) learned on different training sets, compared to other techniques. All global descriptors use GeM pooling \cite{radenovic2018fine}.}
\label{tab:training_set_comparison}
\end{table}

\subsection{Comparing Training Datasets}
We assess the utility of our dataset's training split for transfer learning, by using it to learn global descriptor models and evaluating them on independent landmark retrieval datasets: Revisited Oxford ($\mathcal{R}$Oxf) and Revisited Paris ($\mathcal{R}$Par) \cite{radenovic2018revisiting}.
A ResNet-101 \cite{he2015deep} model is used, with GeM \cite{radenovic2018fine} pooling, trained with ArcFace loss \cite{deng2019arcface}.
Results are presented in \tabref{tab:training_set_comparison}, where we compare against models trained on other datasets, as well as recent state-of-the-art results -- including methods based on global descriptors \cite{radenovic2018fine,revaud2019aploss}, local feature aggregation \cite{radenovic2018revisiting,teichmann2019d2r} and unified global+local features \cite{Cao2020DELG}.
Note that ``SP'' denotes methods using local feature-based spatial verification for re-ranking.

Model training on GLDv2-train-clean provides a substantial boost in performance, compared to training on GLDv1-train: mean average precision (mAP) improves by up to $10$\%.
We also compare with models trained on the Landmarks-full and Landmarks-clean datasets \cite{gordo2016deep}: performance is significantly lower, which is likely due to their much smaller scale.
Our simple global descriptor baseline even outperforms all methods on the $\mathcal{R}$Par dataset, and comes close to the state-of-the-art in $\mathcal{R}$Oxf.
Results in the GLDv2-train-no-tail variant show high performance, although a little lower than GLDv2-train-clean in all cases.

\subsection{Benchmarking}

\begin{table}[t]
\setlength{\tabcolsep}{2pt}
\footnotesize
\centering
\begin{tabular}{l c c c}
Technique & Training Dataset & Testing & Validation \\
\midrule
\multirow{4}{*}{ResNet101+ArcFace} & Landmarks-full \cite{gordo2016deep} & $23.20$ & $20.07$ \\
 & Landmarks-clean \cite{gordo2016deep} & $22.23$ & $20.48$ \\
 & GLDv1-train \cite{noh2017large} & $33.25$ & $33.21$ \\
 & GLDv2-train-clean & $28.56$ & $26.89$ \\
\midrule
DELF-KD-tree \cite{noh2017large} & \multirow{7}{*}{GLDv1-train \cite{noh2017large}} & $44.84$ & $41.07$\\
DELF-ASMK* \cite{radenovic2018revisiting} & & $16.79$ & -- \\
DELF-ASMK*+SP \cite{radenovic2018revisiting} & & $\mathbf{60.16}$ & -- \\
DELF-R-ASMK* \cite{teichmann2019d2r} & & $47.54$ & -- \\
DELF-R-ASMK*+SP \cite{teichmann2019d2r} & & $54.03$ & -- \\
DELG global-only \cite{Cao2020DELG} & & $32.03$ & $32.52$ \\
DELG global+SP \cite{Cao2020DELG} & & $58.45$ & $56.39$ \\
\end{tabular}
\vspace{-7pt}
\caption{Baseline results (\% $\mu$AP) for the GLDv2 recognition task.}
\label{tab:baselines_recognition}
\end{table}

\begin{table}[t]
\setlength{\tabcolsep}{2pt}
\footnotesize
\centering
\begin{tabular}{l c c c}
Technique & Training Dataset & Testing & Validation \\
\midrule
\multirow{4}{*}{ResNet101+ArcFace} & Landmarks-full \cite{gordo2016deep} & $13.27$ & $10.75$ \\
 & Landmarks-clean \cite{gordo2016deep} & $13.55$ & $11.95$ \\
 & GLDv1-train \cite{noh2017large} & $20.67$ & $18.82$ \\
 & GLDv2-train-clean & $\mathbf{25.57}$ & $\mathbf{23.30}$ \\
\midrule
DELF-ASMK* \cite{radenovic2018revisiting} & \multirow{6}{*}{GLDv1-train \cite{noh2017large}}  & $14.76$ & $13.07$ \\
DELF-ASMK*+SP \cite{radenovic2018revisiting} & & $16.92$ & $15.05$ \\
DELF-R-ASMK* \cite{teichmann2019d2r} & & $18.21$ & $16.32$ \\
DELF-R-ASMK*+SP \cite{teichmann2019d2r} & & $18.78$ & $17.38$ \\
DELG global-only \cite{Cao2020DELG} & & $21.71$ & $20.19$ \\
DELG global+SP \cite{Cao2020DELG} & & $24.54$ & $21.52$ \\
\midrule
ResNet101+AP \cite{revaud2019aploss} &  \multirow{3}{*}{GLDv1-train \cite{noh2017large}} & $18.71$ & $16.30$ \\
ResNet101+Triplet \cite{Weinberger2006TripletLoss} & & $18.94$ & $17.14$ \\
ResNet101+CosFace \cite{wang2018cosface} & & $21.35$ & $18.41$ \\
\end{tabular}
\vspace{-7pt}
\caption{Baseline results (\% mAP@100) for the GLDv2 retrieval task. The bottom three results were reported in \cite{Yokoo20Landmark}.}
\label{tab:baselines_retrieval}.
\end{table}

\begin{table*}[ht]
\footnotesize
\centering
\begin{tabular}{l l r r r r}
 \multirow{2}{*}{Team Name} & \multirow{2}{*}{Technique} & \multicolumn{2}{c}{} & \multicolumn{2}{c}{Before re-annotation} \\ 
 & & \multicolumn{1}{c}{Testing} & \multicolumn{1}{c}{Validation} & \multicolumn{1}{c}{Testing} & \multicolumn{1}{c}{Validation} \\
\midrule
smlyaka \cite{Yokoo20Landmark} & GF ensemble $\rightarrow$ LF $\rightarrow$ category filter & $69.39$ & $65.85$ & $35.54$ & $30.96$ \\
JL \cite{Gu2019Landmark} & GF ensemble $\rightarrow$ LF $\rightarrow$ non-landmark filter & $66.53$ & $61.86$ & $37.61$ & $32.10$ \\
GLRunner \cite{Chen2019Landmark} & GF $\rightarrow$ non-landmark detector $\rightarrow$ GF+classifier & $53.08$ & $52.07$ & $35.99$ & $37.14$ \\
\end{tabular}
\vspace{-7pt}
\caption{Top 3 results on recognition challenge (\% $\mu$AP). GF = global feature similarity search; LF = local feature matching re-ranking.}
\label{tab:recognition_challenge_results}
\end{table*}

\begin{table*}[ht]
\footnotesize
\centering
\begin{tabular}{l l r r r r r r r r}
 \multirow{2}{*}{Team Name} & \multirow{2}{*}{Technique}  & \multicolumn{2}{c}{} & \multicolumn{2}{c}{Before re-annotation} & \multicolumn{2}{c}{Precision@100} \\ 
& & \multicolumn{1}{c}{Testing} & \multicolumn{1}{c}{Validation} & \multicolumn{1}{c}{Testing} & \multicolumn{1}{c}{Validation} &  \multicolumn{1}{c}{After} &  \multicolumn{1}{c}{Before} \\
\midrule
smlyaka \cite{Yokoo20Landmark} & GF ensemble $\rightarrow$ DBA/QE $\rightarrow$ C & $37.19$ & $35.69$ & $37.25$ & $35.68$ & $6.09$ & $4.73$ \\
GLRunner \cite{Chen2019Landmark} & GF ensemble $\rightarrow$ LF $\rightarrow$ DBA/QE $\rightarrow$ C & $34.38$ & $32.04$ & $34.75$ & $32.09$ & $6.42$ & $4.83$ \\
Layer 6 AI \cite{Chang2019Landmark} & GF ensemble $\rightarrow$ LF $\rightarrow$ QE $\rightarrow$ EGT & $32.10$ & $29.92$ & $32.18$ & $29.64$ & $5.13$ & $3.97$ \\
\end{tabular}
\vspace{-7pt}
\caption{Top 3 results on retrieval challenge (\% mAP@100). GF = global feature similarity search; LF = local feature matching re-ranking; DBA = database augmentation; QE = query expansion; C = re-ranking based on classifier predictions; EGT = Explore-Exploit Graph Traversal. The last two columns show the effect of the re-annotation on the retrieval precision on the testing set (\% Precision@100).}
\label{tab:retrieval_challenge_results}
\end{table*}

\tabref{tab:baselines_recognition} and \tabref{tab:baselines_retrieval} show results of baseline methods for the recognition and retrieval tasks, respectively. The methods shown use deep local and global features extracted with models that were trained using different datasets and loss functions. All global descriptors use GeM \cite{radenovic2018fine} pooling. For recognition with global descriptors, all methods compose landmark predictions by aggregating the sums of cosine similarities of the top-5 retrieved images; the landmark with the highest sum is used as the predicted label and its sum of cosine similarities is used as the confidence score. For methods with SP, we first find the global descriptor nearest neighbors; then spatially verify the top $100$ images; sort images based on the number of inliers; and aggregate scores over the top-5 images to compose the final prediction, where the score of each image is given by $\frac{min(l,70)}{70} + g$, where $l$ is the number of inliers and $g$ the global descriptor cosine similarity.
For DELF-KD-tree, we use the system proposed in \cite{noh2017large} to obtain the top prediction for each query (if any).

In all cases, training on GLDv1 or GLDv2 improves performance substantially when compared to training on Landmarks-full/clean; for the retrieval task, GLDv2 training performs better, while for the recognition task, GLDv1 performs better.
In the retrieval task, our global descriptor approach trained on GLDv2 outperforms all others; in this case, we also report results from \cite{Yokoo20Landmark} comparing different loss functions; CosFace and ArcFace perform similarly, while Triplet and AP losses perform worse.
In the recognition case, a system purely based on local feature matching with DELF-KD-tree outperforms global descriptors; better performance is obtained when combining local features with global features (DELG), or using local feature aggregation techniques (DELF-ASMK$^\star$+SP).
Note that even better performance may be obtained by improving the combination of local and global scores, as presented in \cite{Cao2020DELG}.

\vspace{-5pt}

\subsection{Challenge Results}
\label{sec:challenge_results}

\tabref{tab:recognition_challenge_results} and \tabref{tab:retrieval_challenge_results} present the top 3 results from the public challenges, for the recognition and retrieval tracks, respectively.
These results are obtained with complex techniques involving ensembling of multiple global and/or local features, usage of trained detectors/classifiers to filter queries, and several query/database expansion techniques.

The most important building block in these systems is the global feature similarity search, which is the first stage in all successful approaches.
These were learned with different backbones such as ResNet \cite{he2015deep}, ResNeXt \cite{xie2017aggregated}, Squeeze-and-Excitation \cite{hu2018squeeze}, FishNet \cite{sun2018fishnet} and Inception-V4 \cite{szegedy2017inception}; pooling methods such as SPoC \cite{babenko2015iccv}, RMAC \cite{tolias2015image} or GeM \cite{radenovic2018fine}; loss functions such as ArcFace \cite{deng2019arcface}, CosFace \cite{wang2018cosface}, N-pairs \cite{sohn2016improved} and triplet \cite{schroff2015facenet}.
Database-side augmentation \cite{arandjelovic2012three} is also often used to improve image representations.

The second most widely used type of method is local feature matching re-ranking, with DELF \cite{noh2017large}, SURF \cite{bay2008speeded} or SIFT \cite{Lowe2004}.
Other re-ranking techniques which are especially important for retrieval tasks, such as query expansion (QE) \cite{chum2011total,radenovic2018fine} and graph traversal \cite{chang2019explore}, were also employed.

These challenge results can be useful as references for future research.
Even with such complex methods, there is still substantial room for improvement in both tasks, indicating that landmark recognition and retrieval are far from solved.

\subsection{Effect of Re-annotation}
The goal of the re-annotation (\secref{sec:reannotation}) was to fill gaps in the ground truth where index images showing the same landmark as a query were not marked as relevant, or where relevant class annotations were missing. To show the effect of this on the metrics, Tab.\ \ref{tab:recognition_challenge_results} and \ref{tab:retrieval_challenge_results} also list the scores of the top methods from the challenge before re-annotation. There is a clear improvement in $\mu$AP for the recognition challenge, which is due to a large number of correctly recognized instances that were previously not counted as correct. However, a similar improvement cannot be observed for the retrieval results. This is because by the design of the the dataset, the retrieval annotations are on the class level rather than the image level. Therefore, if a class is marked as relevant for a query, all of its images are, regardless of whether they have shared content with the query image. So, while the measured \emph{precision} of retrieval increases, the measured \emph{recall} decreases, overall resulting in an almost unchanged mAP score. This is illustrated in the last two columns of \tabref{tab:retrieval_challenge_results}, which shows that Precision@100 consistently increases as an effect of the re-annotation.

\vspace{-5pt}

\section{Conclusion}

\vspace{-2pt}

We have presented the \GLD{}, a new large-scale benchmark for image retrieval and instance recognition. It is the largest such dataset to date and presents several real-world challenges that were not present in previous datasets, such as extreme class imbalance and out-of-domain test images. We hope that the \GLD{} will help advance the state of the art and foster research that deals with these novel challenges for instance recognition and image retrieval.

\PAR{Acknowledgements.}
{
We would like to thank the Wikimedia Foundation and the Wikimedia Commons contributors for the immensely valuable source of image data they created, Kaggle for their support in organizing the challenges, CVDF for hosting the dataset and the co-organizers of the Landmark Recognition workshops at CVPR'18 and CVPR'19. We also thank all teams participating in the Kaggle challenges, especially those whose solutions we used for re-annotation. Special thanks goes to team smlyaka \cite{Yokoo20Landmark} for contributing the cleaned-up dataset and several baseline experiments.}

\section*{Appendix A. Comparison of Retrieval Subset with Oxford and Paris Datasets}
We offer a more detailed comparison of the \textit{retrieval subset} of the \GLD{} (here denoted GLDv2-retrieval) with the $\mathcal{R}$Oxford and $\mathcal{R}$Paris datasets \cite{Philbin07,Philbin2008,radenovic2018revisiting}, which are popular for image retrieval research.

\PAR{Scale.} While the $\mathcal{R}$Oxford and $\mathcal{R}$Paris datasets cover 11 landmarks each and focus on a single city, GLDv2-retrieval has 101k landmarks from all over the world. While the $\mathcal{R}$Oxford and $\mathcal{R}$Paris datasets have 70 query images each, GLDv2-retrieval has 1.1k query images. The $\mathcal{R}$Oxford and $\mathcal{R}$Paris datasets have 5k and 6k index images of landmarks, respectively and additionally have a set of 1M random distractor images, called $\mathcal{R}$1M. Retrieval scores are typically reported both with and without including the distractor set in the index. GLDv2-retrieval has 762k index images of landmarks and has no additional distractors. When including the 1M distractor set, the $\mathcal{R}$Oxford/$\mathcal{R}$Paris index becomes larger than GLDv2-retrieval's index.
However, there is a difference as to how these index images are collected.
For GLDv2-retrieval, index images are collected from an online database with tagged landmarks.
On the other hand, $\mathcal{R}$1M is collected by filtering unconstrained web images with semi-automatic methods, to select those which are the most challenging for recent landmark retrieval techniques; many of them contain actual landmarks, while others may contain images from other domains but which may lead to image representations which ``trick'' recent landmark retrieval techniques.

When not using the distractor set, the $\mathcal{R}$Oxford and $\mathcal{R}$Paris datasets are more accessible when limited resources are available and evaluations on them have much shorter turnaround times. GLDv2-retrieval covers a wider range of landmarks, so we expect results on it to be more representative of practical applications.
Recent papers \cite{tolias2020learning} have also reported results with a small subset of GLDv2, which we believe is a good direction for making evaluations more feasible when only limited resources are available; using the full dataset should be required, though, to draw more robust conclusions.

\PAR{Evaluation protocol.} The query set of GLDv2-retrieval is split into a validation and a testing subset, allowing for a clean evaluation protocol that avoids overfitting: methods should tune performance using the validation split, and only report the testing score for the best tuned version. On the other hand, the $\mathcal{R}$Oxford and $\mathcal{R}$Paris datasets do not offer such a split. In practice, frequent testing during development is often performed without using the distractor set, and experiments with the distractors are done less frequently, to assess large-scale performance \cite{teichmann2019d2r,Cao2020DELG,revaud2019aploss}. Thus, the original $\mathcal{R}$Oxford/$\mathcal{R}$Paris datasets are effectively used as the ``validation" sets and the datasets with distractors are used as the ``testing" sets. This setup is not ideal since the ``validation" set is a subset of the ``testing" set and usually performance on the small scale versions is relatively the same as on the large scale version. This makes it challenging to detect overfitting on these datasets, and in the future we would recommend that a protocol more similar to Tolias \etal \cite{tolias2020learning} would be adopted for these datasets, where a separate validation set is used for tuning.

\PAR{Challenges.} The queries of $\mathcal{R}$Oxford and $\mathcal{R}$Paris are cropped-out regions of images, such as individual windows of a building. These details are often hard to spot in the index images even for humans. The datasets thus pose significant challenges for scale and perspective invariant matching. The queries of GLDv2 are not cropped, so queries can show both the full landmark as well as architectural details. GLDv2 does not explicitly focus on finding small image regions, but provides a natural spectrum of both easy and hard cases for image matching. 

Moreover, index images from $\mathcal{R}$Oxford and $\mathcal{R}$Paris are categorized as ``Easy'', ``Hard'', ``Unclear'' or ``Negative'' for each different query -- leading to different experimental protocols ``Easy'', ``Medium'', ``Hard'', depending on the types of index images expected to be retrieved. In these datasets, the common experimental setup is to report results only for Medium and Hard protocols (as suggested by the main results table in \cite{radenovic2018revisiting}). In contrast, the \GLD{} index images can only be ``Positive'' or ``Negative'', and there is a single protocol. In this way, we believe that our dataset will more accurately capture effects of easy queries, which are very common in real-world systems. As a concrete example of differences we can observe, state-of-the-art methods based on local feature aggregation (\eg, DELF-R-ASMK$^\star$ \cite{teichmann2019d2r}), which excel on $\mathcal{R}$Oxford, do not fare as well on GLDv2, being worse than simple embeddings.

\PAR{Application.} Both $\mathcal{R}$Oxford/$\mathcal{R}$Paris and GLDv2 address instance-level retrieval tasks; however, the ground-truth is constructed differently. In $\mathcal{R}$Oxford/$\mathcal{R}$Paris, relevant index images must depict the same instance and the same view as the query image. In contrast, for GLDv2, any index image associated to the same landmark is considered relevant, even if its view does not overlap with the query image.

In conclusion, both our \GLD{} (retrieval subset) and $\mathcal{R}$Oxford/$\mathcal{R}$Paris have pros and cons, capturing different and complementary aspects of the instance-level retrieval problem. For a comprehensive assessment of instance-level retrieval methods, we would suggest future work to include all of these, to offer a detailed performance analysis across different characteristics.

\section*{Appendix B. Preventing Unintended Methods}
For the Kaggle competition, we had to make certain design choices to prevent ``cheating",
 \ie{} to ensure that participants would only use the images themselves and no metadata attached to the images or found on the web. Therefore, we stripped all images of any metadata like geotags or labels. However, this alone was not sufficient since many images have a ``Creative Commons By" (CC-BY) license which requires attributing the author and publishing the original image URL, which would reveal other information.
We therefore chose to use only images with CC0 or Public Domain licenses for the index set, so we could keep their image URLs secret; the same for the query set, except that in this case we also added images collected by crowdsourcing operators.
For the training set however, the landmark labels needed to be released in order to allow training models. So, we used CC-BY images for the training set and include full attribution information with the dataset.

\section*{Appendix C. Sample Images from the Dataset}
To give a qualitative impression of the dataset, we show a selection of dataset images. We would also like to refer readers to the dataset website, where a web interface is available for exploring images from the dataset.

\subsection*{C.1 Intra-Class Variation in Training Set}
\begin{figure*}
\centering
\begin{subfigure}{\textwidth}
\includegraphics[width=\linewidth]{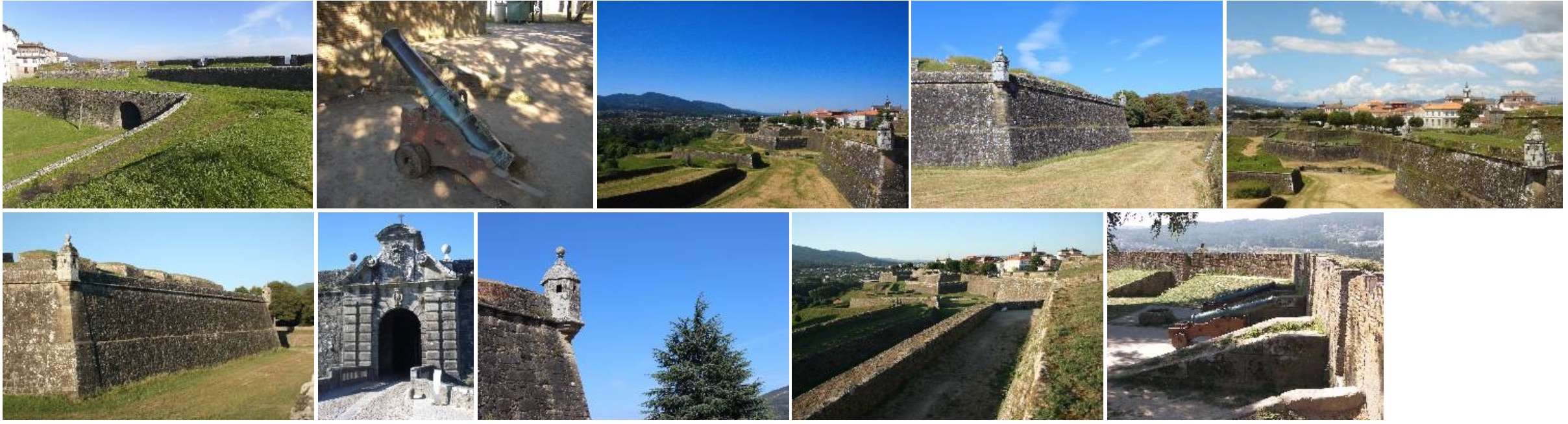}
\caption{Fortifica\c{c}\~{o}es da Pra\c{c}a de Valen\c{c}a do Minho -- Different views of a landmark that covers a large area.}
\end{subfigure}

\begin{subfigure}{\textwidth}
\includegraphics[width=\linewidth]{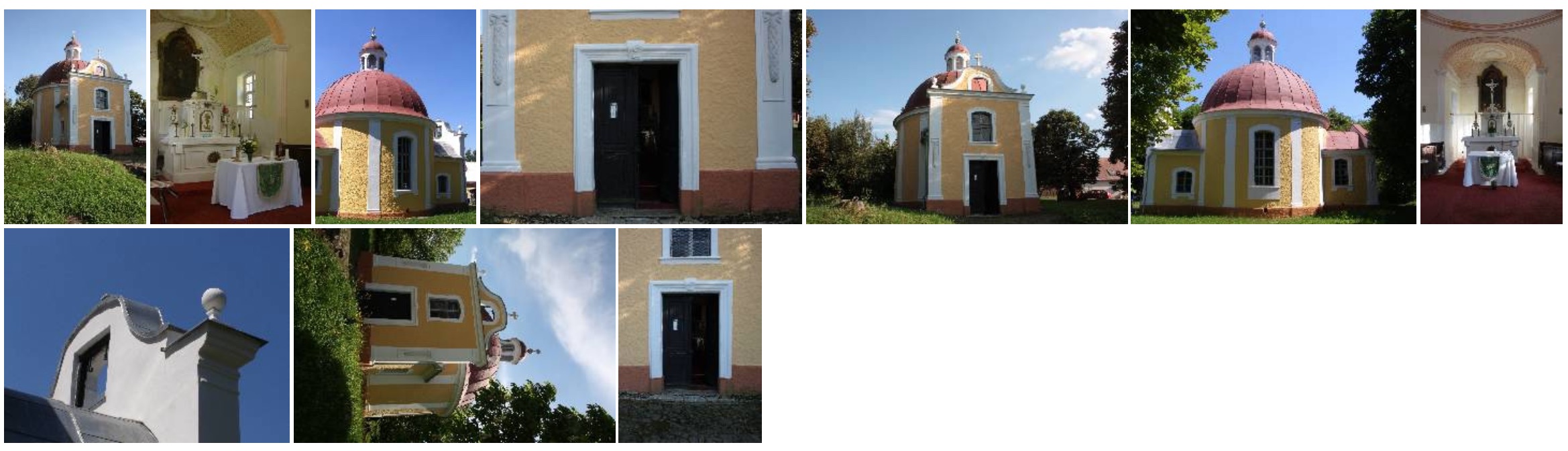}
\caption{Chapel of Saint John of Nepomuk (\v{C}ernousy) -- Inside and outside views as well as details.}
\end{subfigure}

\begin{subfigure}{\textwidth}
\includegraphics[width=\linewidth]{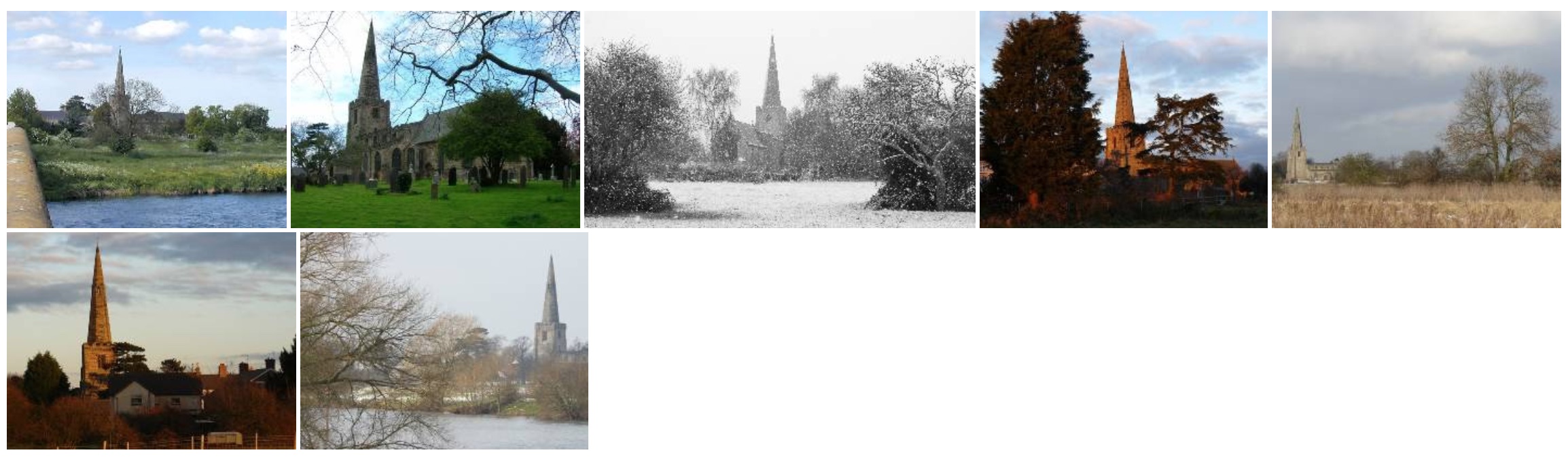}
\caption{All Saints church (Sawley) -- Images showing the landmark in different seasons and under different lighting conditions.}
\end{subfigure}

\begin{subfigure}{\textwidth}
\includegraphics[width=\linewidth]{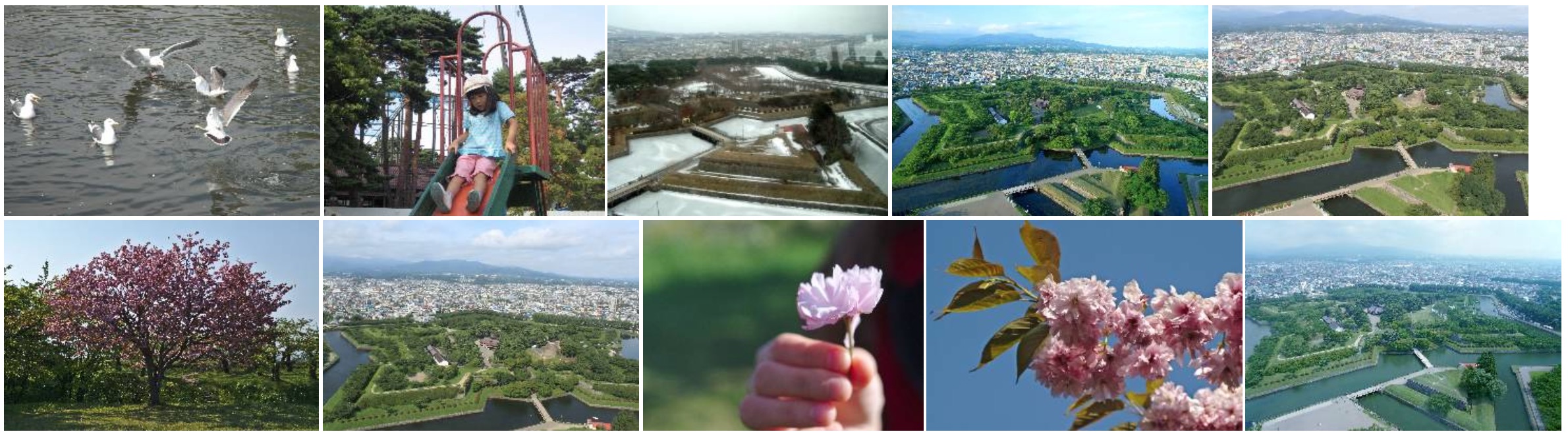}
\caption{Goryokaku (Hakodate) -- Aerial views and details of a park.}
\end{subfigure}

\caption{Sample classes from the training set (1 of 2).}
\label{fig:training_set_1}
\end{figure*}

\begin{figure*}
\centering

\begin{subfigure}{\textwidth}
\includegraphics[width=\linewidth]{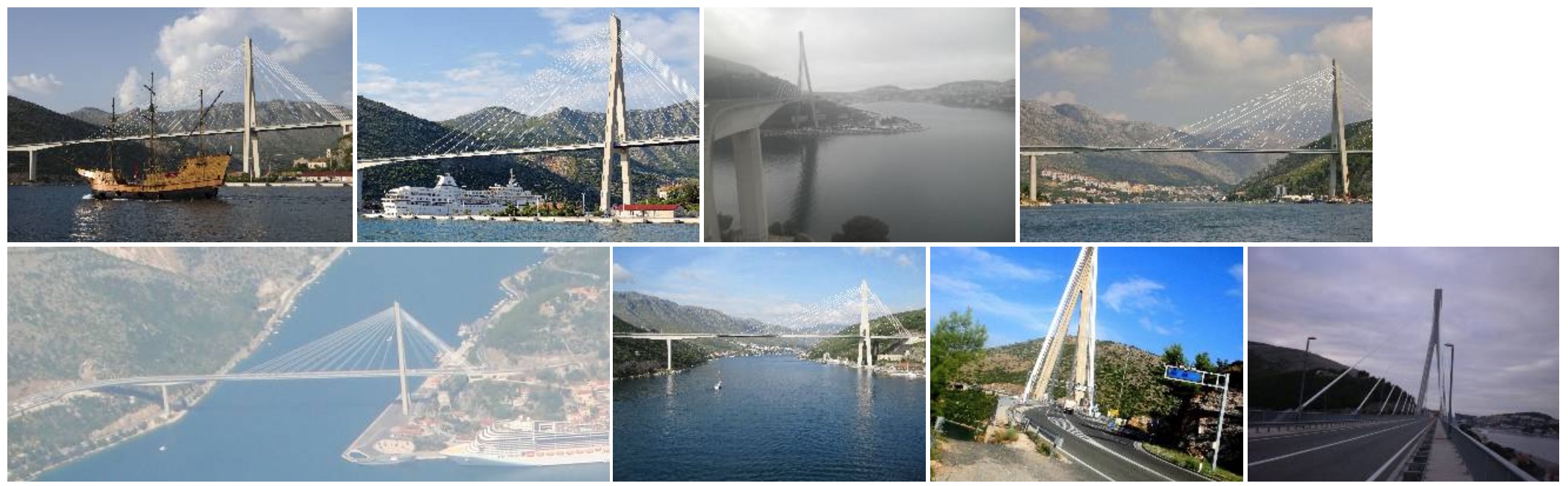}
\caption{Franjo Tudman bridge (Dubrovnik) -- A wide range of views and weather conditions.}
\end{subfigure}

\begin{subfigure}{\textwidth}
\includegraphics[width=\linewidth]{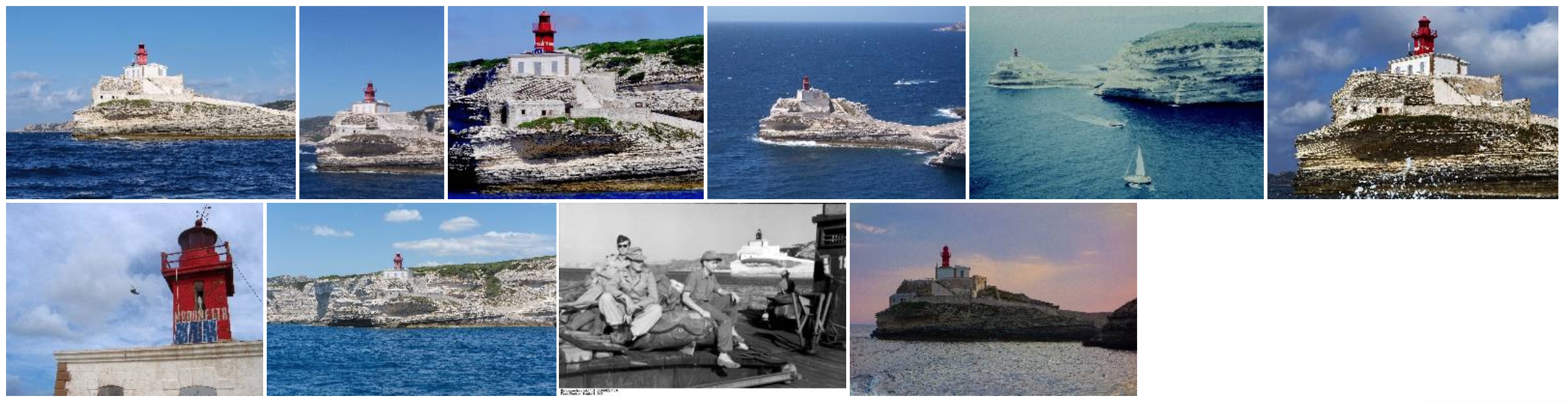}
\caption{Phare de la Madonetta (Bonifacio) -- A wide range of scales and historical photographs.}
\end{subfigure}

\begin{subfigure}{\textwidth}
\includegraphics[width=\linewidth]{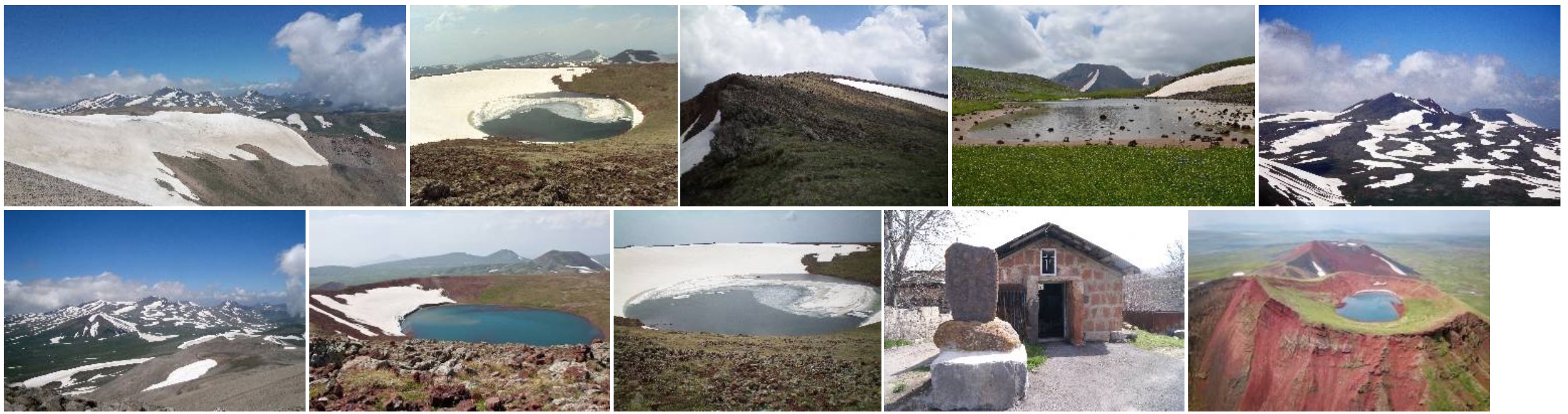}
\caption{Azhdahak (Armenia) -- A natural landmark from different views and with different levels of snow coverage.}
\end{subfigure}

\caption{Sample classes from the training set (2 of 2).}
\label{fig:training_set_2}
\end{figure*}

Figs.\ \ref{fig:training_set_1} and \ref{fig:training_set_2} show a sample of the over \num{200}k classes in the training set. The dataset has a broad coverage of each place, including photos taken from widely different viewing angles, under different lighting and weather conditions and in different seasons. Additionally, it contains historical photos from archives that can help make trained models robust to changes in photo quality and appearance changes over time.

\subsection*{C.2 Retrieval Ground Truth}
\begin{figure*}
    \centering
    \begin{subfigure}{\linewidth}
    \includegraphics[width=1\linewidth]{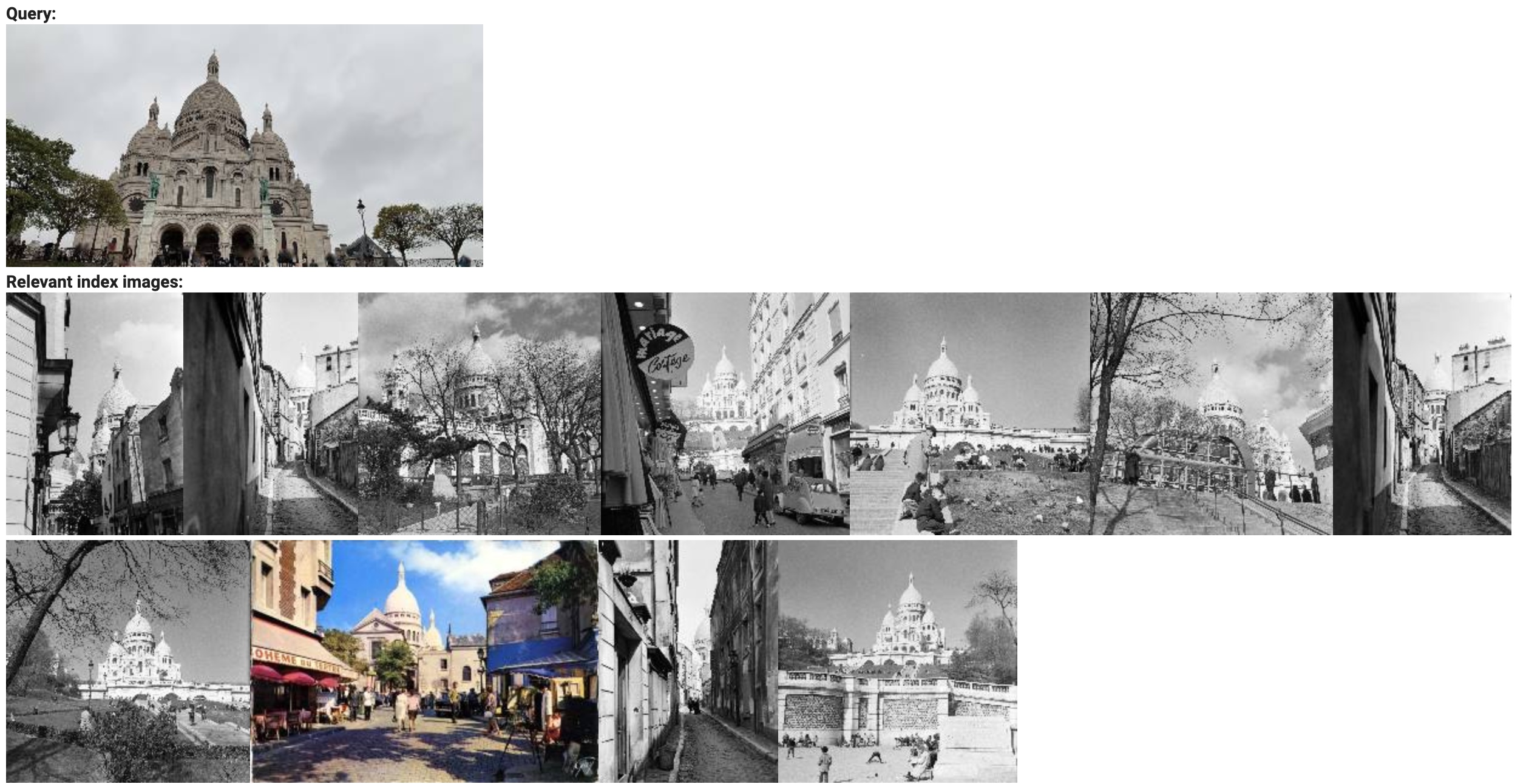}
    \caption{Sacre Coeur -- Different viewpoints and significant occlusion.}
    \end{subfigure}
    \begin{subfigure}{\linewidth}
    \includegraphics[width=1\linewidth]{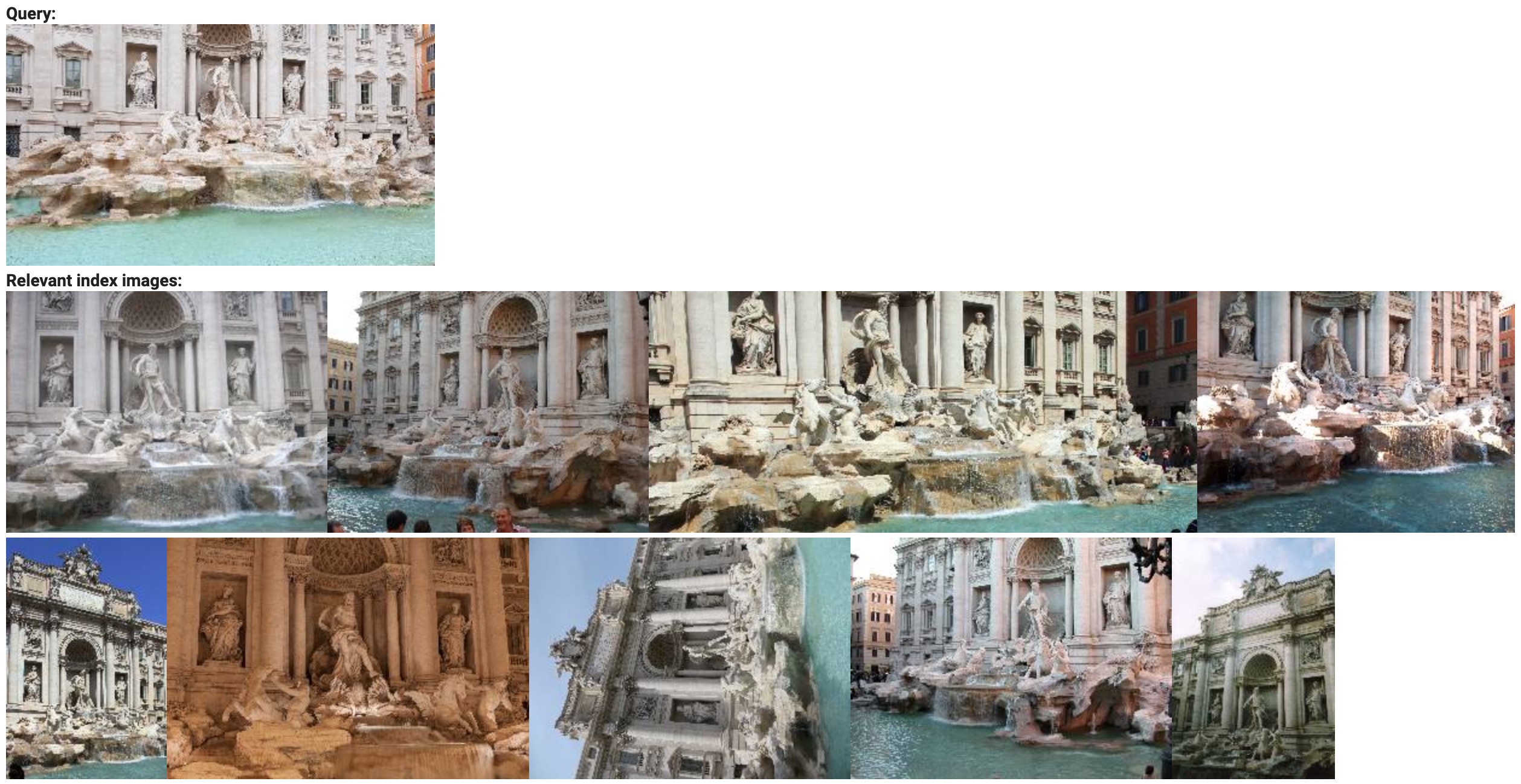}
    \caption{Trevi Fountain -- Different viewpoints and lighting conditions.}
    \end{subfigure}
    \caption{Retrieval task: Query images with a sample of relevant images from the index set (1 of 3).}
    \label{fig:query_index_1}
\end{figure*}

\begin{figure*}
    \begin{subfigure}{\linewidth}
    \includegraphics[width=1\linewidth]{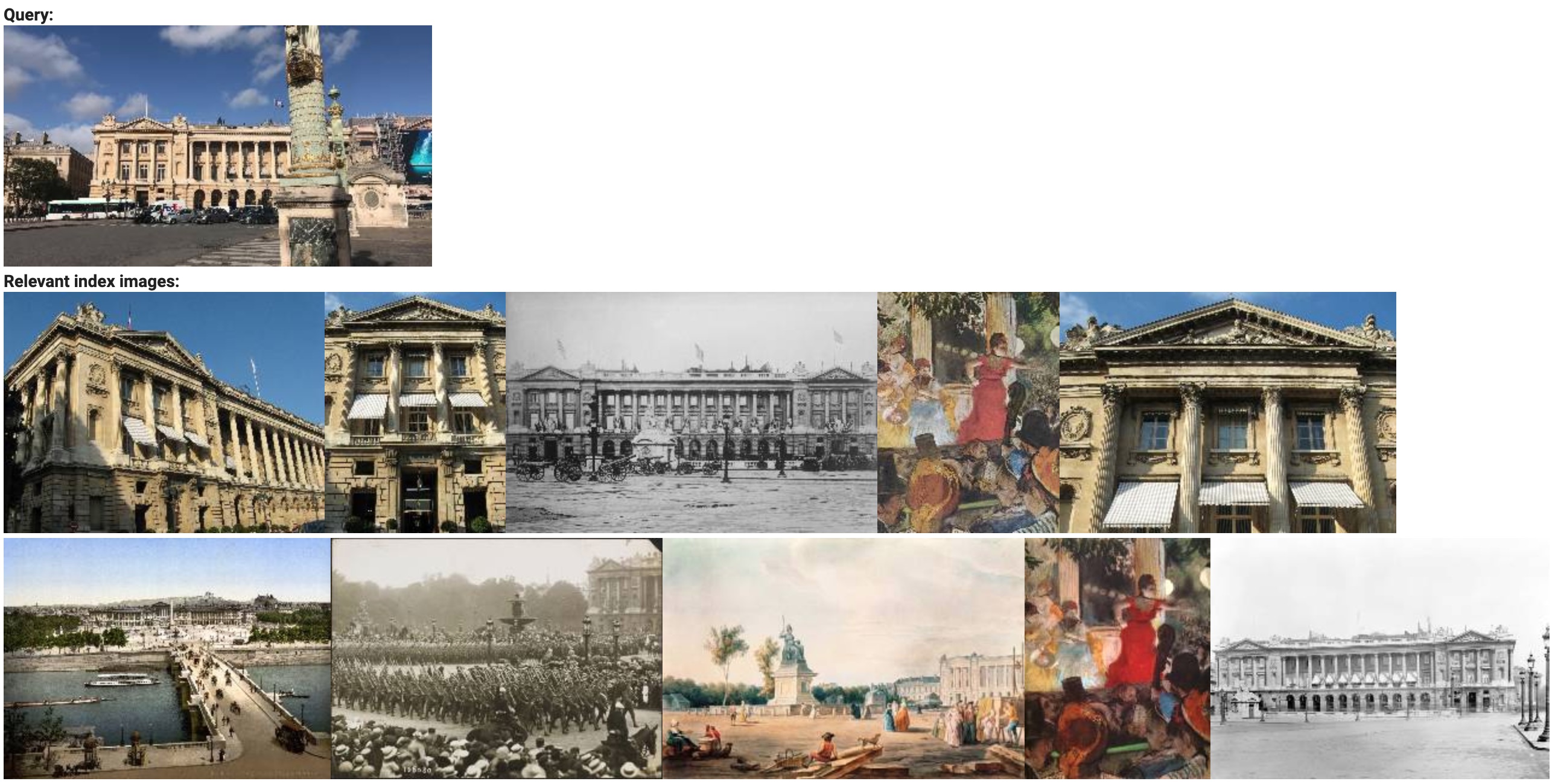}
    \caption{Place de la Concorde -- Significant scale changes, historical photographs and paintings.}
    \end{subfigure}
    \begin{subfigure}{\linewidth}
    \includegraphics[width=1\linewidth]{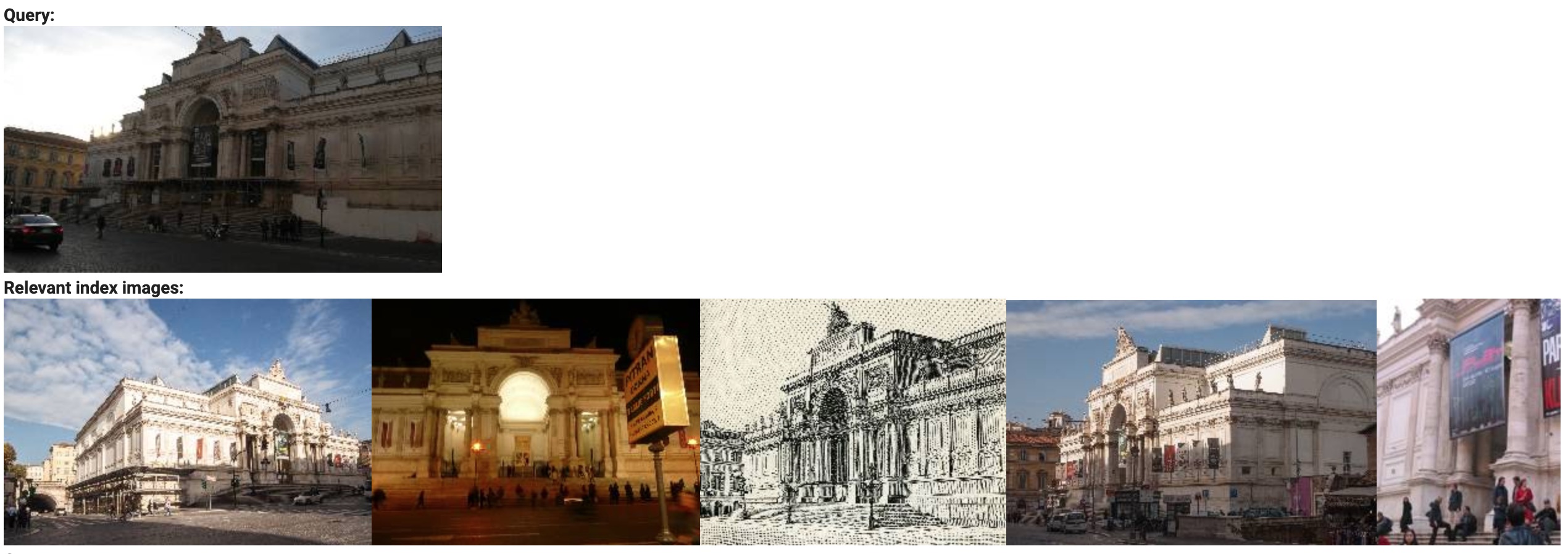}
    \caption{Palazzo delle Esposizioni -- Day and night photos and a monochrome print.}
    \end{subfigure}
    \caption{Retrieval task: Query images with a sample of relevant images from the index set (2 of 3).}
    \label{fig:query_index_2}
\end{figure*}

\begin{figure*}
    \begin{subfigure}{\linewidth}
    \includegraphics[width=1\linewidth]{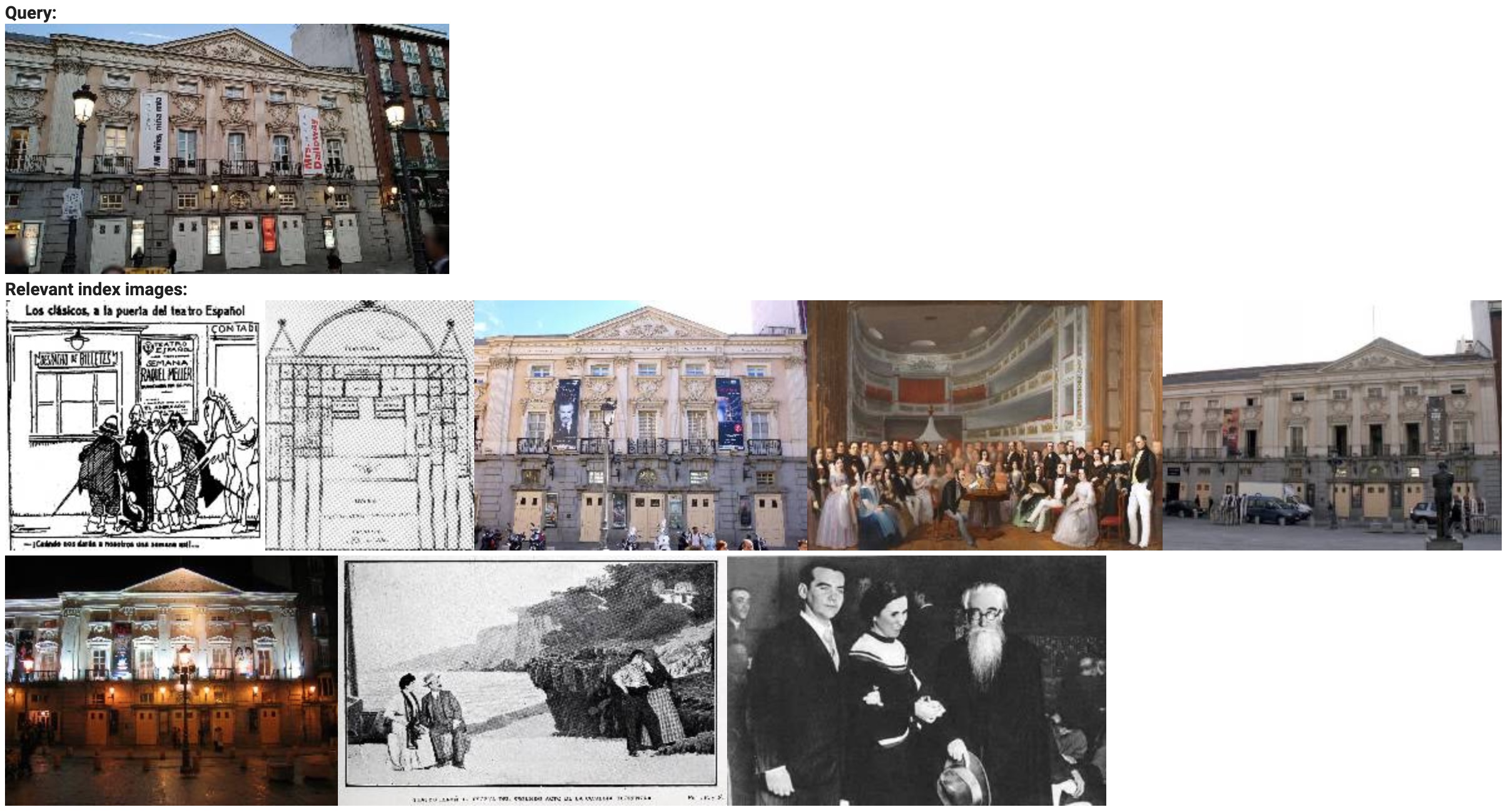}
    \caption{Teatro Espanol -- Some relevant images that are difficult to retrieve: Architectural drawing, painting of the inside, historical photograph of audience members.}
    \end{subfigure}
    \begin{subfigure}{\linewidth}
    \includegraphics[width=1\linewidth]{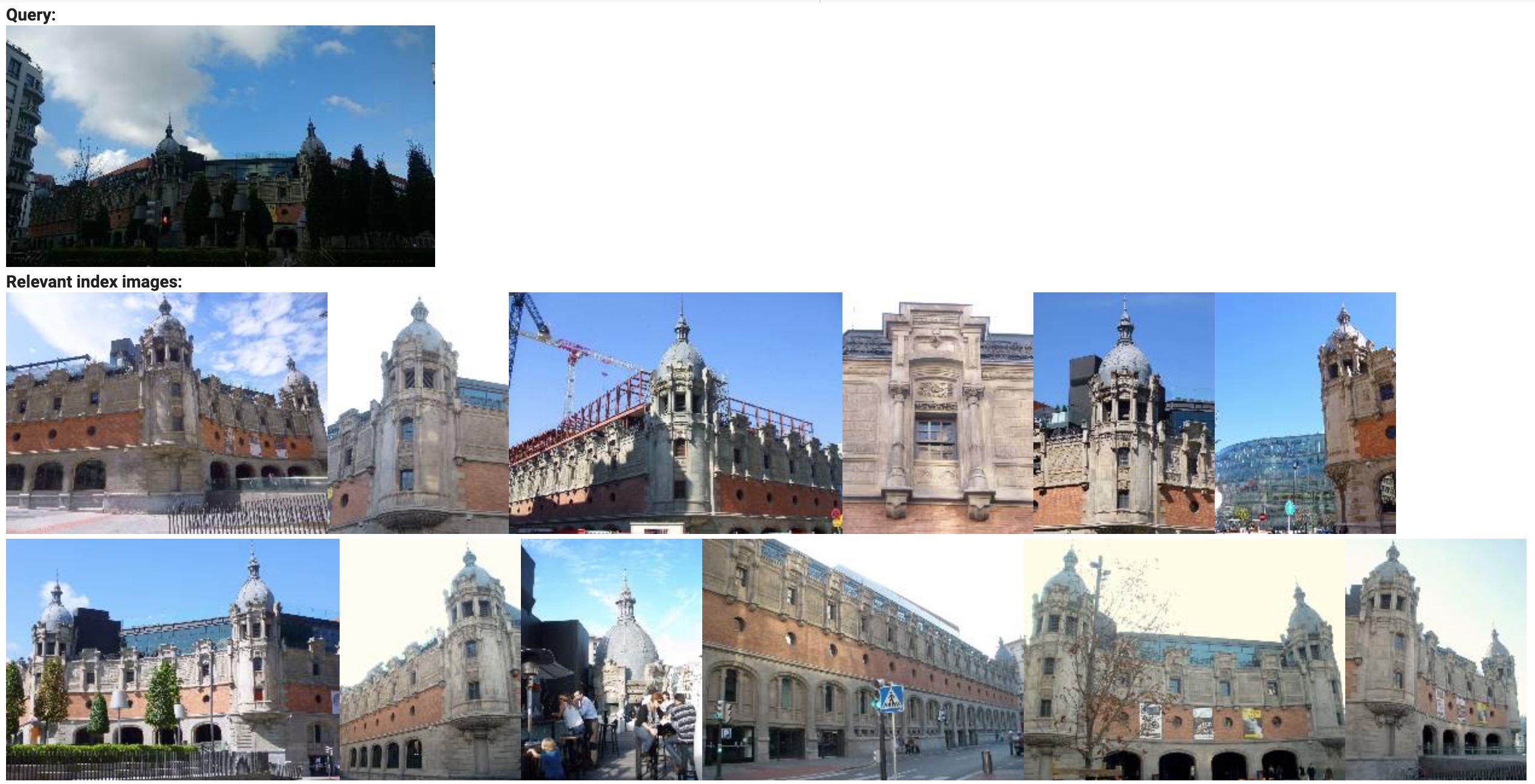}
    \caption{Azkuna Zentroa -- Detail views and photos showing the construction of the building.}
    \end{subfigure}
    \caption{Retrieval task: Query images with a sample of relevant images from the index set (3 of 3).}
    \label{fig:query_index_3}
\end{figure*}

Figs.\ \ref{fig:query_index_1}, \ref{fig:query_index_2} and \ref{fig:query_index_3} show a selection of query images with associated index images, highlighting some of the challenges of the retrieval task. Note that because the retrieval ground truth was created on a class level rather than on the image level, not all relevant index images have shared content with the query image. The retrieval task challenges approaches to be robust to a wide rage of variations, including viewpoint, occlusions, lighting and weather. Moreover, invariance to image domain is required since the index contains both digital and analog photographs as well as some drawings and paintings depicting the landmark.

\subsection*{C.3 Test Set}
\begin{figure*}
\centering
\begin{subfigure}{\textwidth}
\includegraphics[width=.8\linewidth]{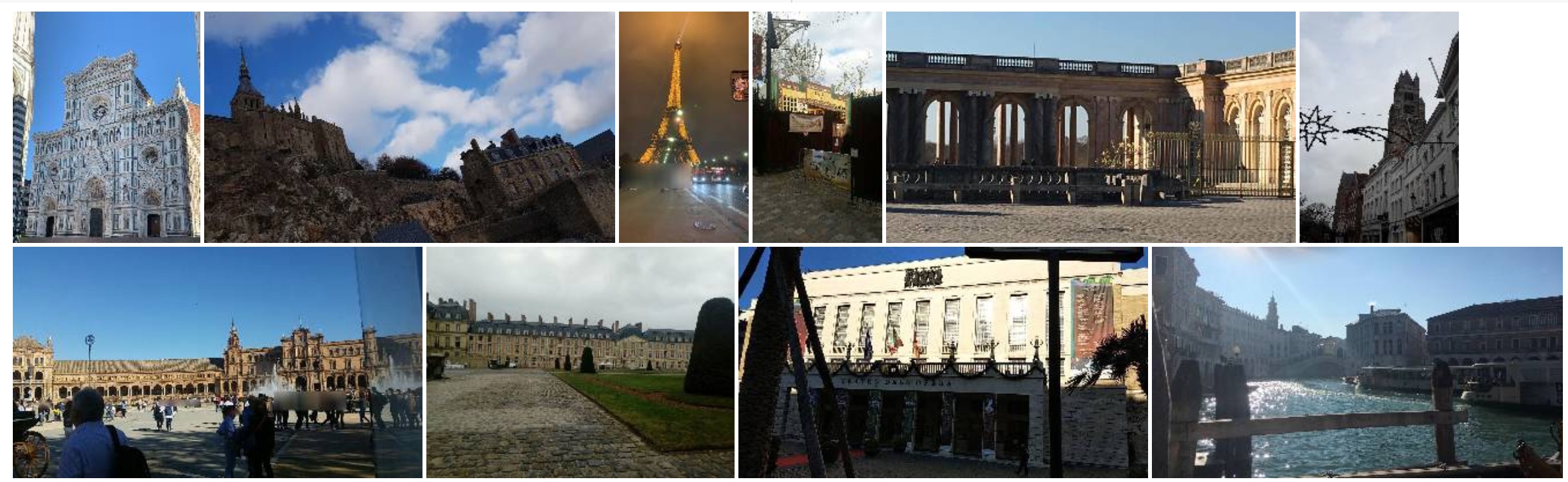}
\caption{Sample landmark images from the test set.}
\label{fig:test_set:positive}
\end{subfigure}
\begin{subfigure}{\textwidth}
\includegraphics[width=\linewidth]{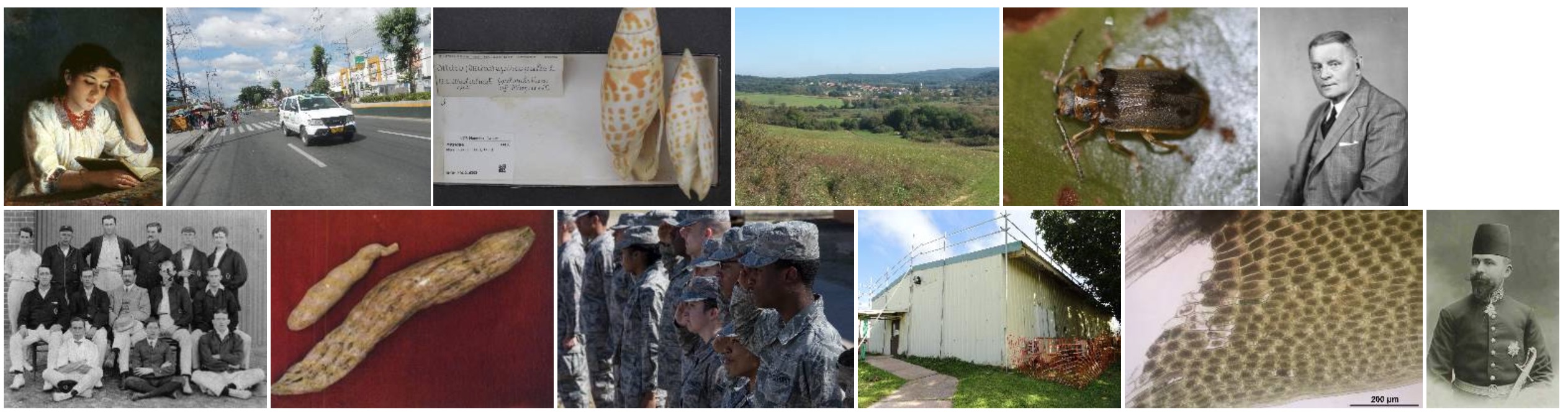}
\includegraphics[width=\linewidth]{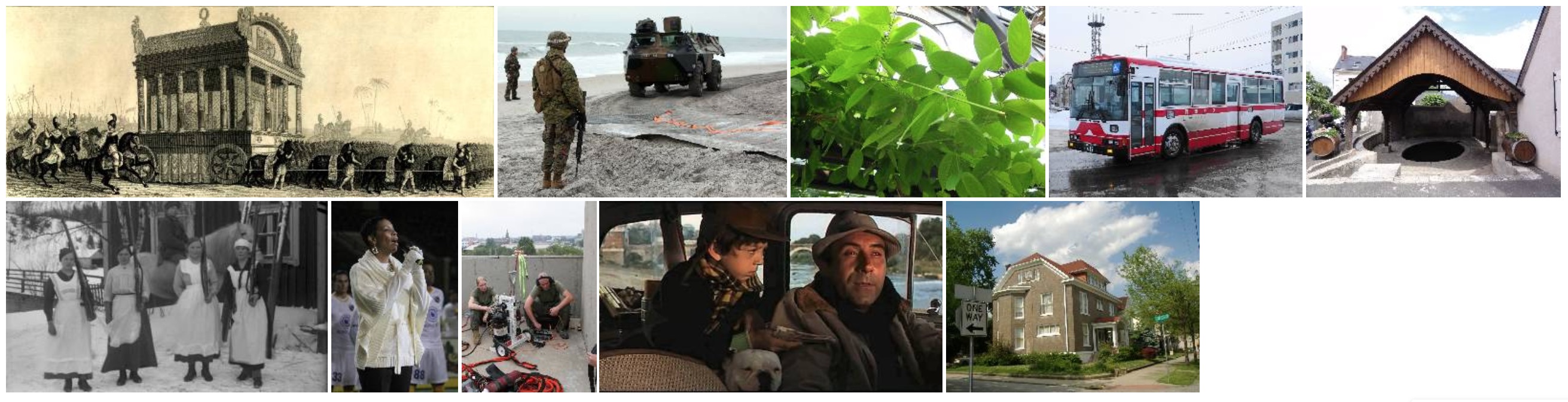}
\includegraphics[width=\linewidth]{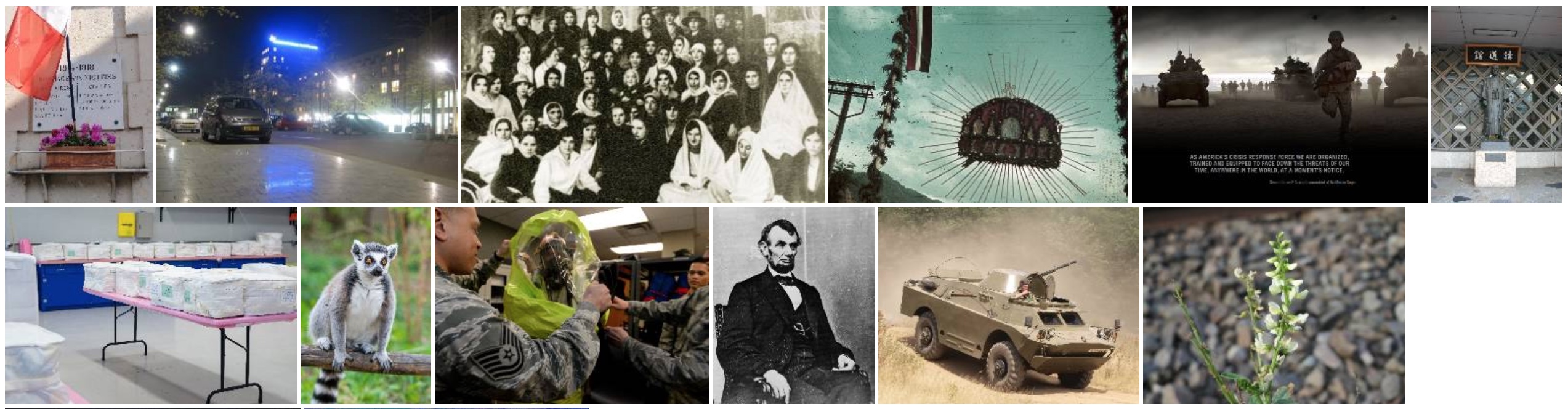}
\caption{Sample out-of-domain images from the test set.}
\label{fig:test_set:negative}
\end{subfigure}
\caption{Sample images from the test set.}
\label{fig:test_set}
\end{figure*}

Fig.\ \ref{fig:test_set} shows images from the test set. The test set consists of \num{1.1}\% images of natural and human-made landmarks, as shown in Fig.\ \ref{fig:test_set:positive}. These images were taken with smartphones by crowdsourcing operators. They therefore represent realistic query images to visual recognition applications. Fig.\ \ref{fig:test_set:negative} shows a sample of the \num{98.9}\% out-of-domain images in the test set that were collected from Wikimedia Commons. Note that a small fraction of test set images showing landmarks do not have ground truth labels since their landmarks do not exist in the training or query sets.

\clearpage

\afterpage{
{\small
\bibliographystyle{ieee_fullname}
\bibliography{literature/gldv2}

\begin{thebibliography}{10}\itemsep=-1pt

\bibitem{agarwal2011building}
S. Agarwal, Y. Furukawa, N. Snavely, I. Simon, B. Curless, S. Seitz, and R.
  Szeliski.
\newblock {Building Rome in a Day}.
\newblock {\em Communications of the ACM}, 2011.

\bibitem{arandjelovic2011smooth}
R. Arandjelovic and A. Zisserman.
\newblock {Smooth object retrieval using a bag of boundaries}.
\newblock In {\em Proc. ICCV}, 2011.

\bibitem{arandjelovic2012three}
R. Arandjelovic and A. Zisserman.
\newblock {Three Things Everyone Should Know to Improve Object Retrieval}.
\newblock In {\em Proc. CVPR}, 2012.

\bibitem{avrithis2010retrieving}
Y. Avrithis, Y. Kalantidis, G. Tolias, and E. Spyrou.
\newblock {Retrieving Landmark and Non-landmark Images from Community Photo
  Collections}.
\newblock In {\em Proc. ACM MM}, 2010.

\bibitem{avrithis2010feature}
Y. Avrithis, G. Tolias, and Y. Kalantidis.
\newblock {Feature Map Hashing: Sub-linear Indexing of Appearance and Global
  Geometry}.
\newblock In {\em Proc. ACM MM}, 2010.

\bibitem{babenko2015iccv}
A. Babenko and V. Lempitsky.
\newblock {Aggregating Local Deep Features for Image Retrieval}.
\newblock In {\em Proc. ICCV}, 2015.

\bibitem{babenko2014neural}
A. Babenko, A. Slesarev, A. Chigorin, and V. Lempitsky.
\newblock {Neural Codes for Image Retrieval}.
\newblock In {\em Proc. ECCV}, 2014.

\bibitem{bai2018group}
Y. Bai, Y. Lou, F. Gao, S. Wang, Y. Wu, and L. Duan.
\newblock {Group-Sensitive Triplet Embedding for Vehicle Reidentification}.
\newblock {\em IEEE Transactions on Multimedia}, 2018.

\bibitem{bay2008speeded}
H. Bay, A. Ess, T. Tuytelaars, and L. Van~Gool.
\newblock {Speeded-Up Robust Features (SURF)}.
\newblock {\em Computer Vision and Image Understanding}, 2008.

\bibitem{Cao2020DELG}
B. Cao, A. Araujo, and J. Sim.
\newblock {Unifying Deep Local and Global Features for Image Search}.
\newblock In {\em Proc. ECCV}, 2020.

\bibitem{chandrasekhar2011the}
V. Chandrasekhar, D. Chen, S.~S. Tsai, N.~M. Cheung, H. Chen, G. Takacs, Y.
  Reznik, R. Vedantham, R. Grzeszczuk, J. Bach, and B. Girod.
\newblock {The Stanford Mobile Visual Search Dataset}.
\newblock In {\em Proc. ACM Multimedia Systems Conference}, 2011.

\bibitem{Chang2019Landmark}
C. Chang, H. Rai, S.~K. Gorti, J. Ma, C. Liu, G. Yu, and M. Volkovs.
\newblock {Semi-Supervised Exploration in Image Retrieval}.
\newblock {\em arXiv:1906.04944}, 2019.

\bibitem{chang2019explore}
C. Chang, G. Yu, C. Liu, and M. Volkovs.
\newblock {Explore-Exploit Graph Traversal for Image Retrieval}.
\newblock In {\em Proc. CVPR}, 2019.

\bibitem{chen2011city}
D. Chen, G. Baatz, K. Koser, S. Tsai, R. Vedantham, T. Pylvanainen, K. Roimela,
  X. Chen, J. Bach, M. Pollefeys, B. Girod, and R. Grzeszczuk.
\newblock {City-Scale Landmark Identification on Mobile Devices}.
\newblock In {\em Proc. CVPR}, 2011.

\bibitem{Chen2019Landmark}
K. Chen, C. Cui, Y. Du, X. Meng, and H. Ren.
\newblock {2nd Place and 2nd Place Solution to Kaggle Landmark Recognition and
  Retrieval Competition 2019}.
\newblock {\em arXiv:1906.03990}, 2019.

\bibitem{DelChiaro2019noisyart}
R. Chiaro, A. Bagdanov, and A. Bimbo.
\newblock {{\{}NoisyArt{\}}: A Dataset for Webly-supervised Artwork
  Recognition}.
\newblock In {\em Proc. VISAPP}, 2019.

\bibitem{chum2011total}
O. Chum, A. Mikulik, M. Perdoch, and J. Matas.
\newblock {Total Recall II: Query Expansion Revisited}.
\newblock In {\em Proc. CVPR}, 2011.

\bibitem{deng2019arcface}
J. Deng, J. Guo, N. Xue, and S. Zafeiriou.
\newblock {ArcFace: Additive Angular Margin Loss for Deep Face Recognition}.
\newblock In {\em Proc. CVPR}, 2019.

\bibitem{fehervari2019scalable}
I. Fehervari and S. Appalaraju.
\newblock {Scalable Logo Recognition using Proxies}.
\newblock In {\em Proc. WACV}, 2019.

\bibitem{feifei2004learning}
L. Fei-Fei, R. Fergus, and P. Perona.
\newblock {Learning Generative Visual Models from Few Training Examples: An
  Incremental Bayesian Approach Tested on 101 Object Categories}.
\newblock In {\em Proc. CVPR Workshops}, 2004.

\bibitem{ge2019deepfashion2}
Y. Ge, R. Zhang, L. Wu, X. Wang, X. Tang, and P. Luo.
\newblock {DeepFashion2: A Versatile Benchmark for Detection, Pose Estimation,
  Segmentation and Re-Identification of Clothing Images}.
\newblock In {\em Proc. CVPR}, 2019.

\bibitem{gordo2016deep}
A. Gordo, J. Almazan, J. Revaud, and D. Larlus.
\newblock {Deep Image Retrieval: Learning Global Representations for Image
  Search}.
\newblock In {\em Proc. ECCV}, 2016.

\bibitem{griffin2007}
G. Griffin, A. Holub, and P. Perona.
\newblock {Caltech-256 Object Category Dataset}.
\newblock Technical Report 7694, California Institute of Technology, 2007.

\bibitem{Gu2019Landmark}
Y. Gu and C Li.
\newblock {Team JL Solution to Google Landmark Recognition 2019}.
\newblock {\em arXiv:1906.11874}, 2019.

\bibitem{he2015deep}
K. He, X. Zhang, S. Ren, and J. Sun.
\newblock {Deep Residual Learning for Image Recognition}.
\newblock In {\em Proc. CVPR}, 2016.

\bibitem{vanhorn2018the}
G.V. Horn, O.M. Aodha, Y. Song, Y. Cui, C. Sun, A. Shepard, H. Adam, P. Perona,
  and S. Belongie.
\newblock {The iNaturalist Species Classification and Detection Dataset}.
\newblock In {\em Proc. CVPR}, 2018.

\bibitem{hu2018squeeze}
J. Hu, L. Shen, and G. Sun.
\newblock {Squeeze-and-Excitation Networks}.
\newblock In {\em Proc. CVPR}, 2018.

\bibitem{Jegou2008}
H. J\'{e}gou, M. Douze, and C. Schmid.
\newblock {Hamming Embedding and Weak Geometric Consistency for Large Scale
  Image Search}.
\newblock In {\em Proc. ECCV}, 2008.

\bibitem{jegou2012aggregating}
H. J\'{e}gou, F. Perronnin, M. Douze, J. Sanchez, P. Perez, and C. Schmid.
\newblock {Aggregating Local Image Descriptors into Compact Codes}.
\newblock {\em IEEE Transactions on Pattern Analysis and Machine Intelligence},
  2012.

\bibitem{joly2009logo}
A. Joly and O. Buisson.
\newblock {Logo Retrieval with a Contrario Visual Query Expansion}.
\newblock In {\em Proc. ACM MM}, 2009.

\bibitem{kalantidis2011scalable}
Y. Kalantidis, LG. Pueyo, M. Trevisiol, R. van Zwol, and Y. Avrithis.
\newblock {Scalable Triangulation-based Logo Recognition}.
\newblock In {\em Proc. ICMR}, 2011.

\bibitem{knopp2010avoiding}
J. Knopp, J. Sivic, and T. Pajdla.
\newblock {Avoiding Confusing Features in Place Recognition}.
\newblock In {\em Proc. ECCV}, 2010.

\bibitem{krause20133d}
J. Krause, M. Stark, J. Deng, , and L. Fei-Fei.
\newblock {3d Object Representations for Fine-Grained Categorization}.
\newblock In {\em Proc. ICCV Workshops}, 2013.

\bibitem{kuznetsova2018open}
A. Kuznetsova, H. Rom, N. Alldrin, J. Uijlings, I. Krasin, J. Pont-Tuset, S.
  Kamali, S. Popov, M. Malloci, T. Duerig, and V. Ferrari.
\newblock {The Open Images Dataset V4: Unified Image Classification, Object
  Detection, and Visual Relationship Detection at Scale}.
\newblock {\em arXiv:1811.00982}, 2018.

\bibitem{li2012worldwide}
Y. Li, N. Snavely, D. Huttenlocher, and P. Fua.
\newblock {Worldwide Pose Estimation using 3D Point Clouds}.
\newblock In {\em Proc. ECCV}, 2012.

\bibitem{lin2014coco}
T. Lin, M. Maire, S. Belongie, L. Bourdev, R. Girshick, J. Hays, P. Perona, D.
  Ramanan, P. Doll{\'{a}}r, and C. Zitnick.
\newblock Microsoft {COCO:} common objects in context.
\newblock In {\em Proc. ECCV}, 2014.

\bibitem{liu2016deepfashion}
Z. Liu, P. Luo, S. Qiu, X. Wang, and X. Tang.
\newblock {DeepFashion: Powering Robust Clothes Recognition and Retrieval with
  Rich Annotations}.
\newblock In {\em Proc. CVPR}, 2016.

\bibitem{Lowe2004}
D. Lowe.
\newblock {Distinctive Image Features from Scale-Invariant Keypoints}.
\newblock {\em IJCV}, 2004.

\bibitem{noh2017large}
H. Noh, A. Araujo, J. Sim, T. Weyand, and B. Han.
\newblock {Large-Scale Image Retrieval with Attentive Deep Local Features}.
\newblock In {\em Proc. ICCV}, 2017.

\bibitem{Perronnin09CVPR}
F. Perronnin, Y. Liu, and J. Renders.
\newblock {A Family of Contextual Measures of Similarity between Distributions
  with Application to Image Retrieval}.
\newblock In {\em Proc. CVPR}, 2009.

\bibitem{Philbin07}
J. Philbin, O. Chum, M. Isard, J. Sivic, and A. Zisserman.
\newblock {Object Retrieval with Large Vocabularies and Fast Spatial Matching}.
\newblock In {\em Proc. CVPR}, 2007.

\bibitem{Philbin2008}
J. Philbin, O. Chum, M. Isard, J. Sivic, and A. Zisserman.
\newblock {Lost in Quantization: Improving Particular Object Retrieval in Large
  Scale Image Databases}.
\newblock In {\em Proc. CVPR}, 2008.

\bibitem{radenovic2018revisiting}
F. Radenovi{\'c}, A. Iscen, G. Tolias, Y. Avrithis, and O. Chum.
\newblock {Revisiting Oxford and Paris: Large-Scale Image Retrieval
  Benchmarking}.
\newblock In {\em Proc. CVPR}, 2018.

\bibitem{radenovic2018fine}
F. Radenovi{\'c}, G. Tolias, and O. Chum.
\newblock {Fine-tuning CNN Image Retrieval with No Human Annotation}.
\newblock {\em IEEE Transactions on Pattern Analysis and Machine Intelligence},
  2018.

\bibitem{revaud2019aploss}
J. Revaud, J. Almazan, R. Rezende, and C.~R. de Souza.
\newblock {Learning with Average Precision: Training Image Retrieval with a
  Listwise Loss}.
\newblock In {\em Proc. CVPR}, 2019.

\bibitem{romberg2011scalable}
S. Romberg, L. Pueyo, R. Lienhart, and R. van Zwol.
\newblock {Scalable Logo Recognition in Real-world Images}.
\newblock In {\em Proc. ICMR}, 2011.

\bibitem{russakovsky2015imagenet}
O. Russakovsky, J. Deng, H. Su, J. Krause, S. Satheesh, S. Ma, Z. Huang, A.
  Karpathy, A. Khosla, M. Bernstein, et~al.
\newblock {ImageNet Large Scale Visual Recognition Challenge}.
\newblock {\em IJCV}, 2015.

\bibitem{schroff2015facenet}
F. Schroff, D. Kalenichenko, and J. Philbin.
\newblock {A Unified Embedding for Face Recognition and Clustering}.
\newblock In {\em Proc. CVPR}, 2015.

\bibitem{sohn2016improved}
K. Sohn.
\newblock {Improved Deep Metric Learning with Multiclass N-Pair Loss
  Objective}.
\newblock In {\em Proc. NIPS}, 2016.

\bibitem{song2016deep}
H. Song, Y. Xiang, S. Jegelka, and S. Savarese.
\newblock {Deep Metric Learning via Lifted Structured Feature Embedding}.
\newblock In {\em Proc. CVPR}, 2016.

\bibitem{sun2018fishnet}
S. Sun, J. Pang, J. Shi, S. Yi, and W. Ouyang.
\newblock {Fishnet: A Versatile Backbone for Image, Region, and Pixel Level
  Prediction}.
\newblock In {\em Proc. NeurIPS}, 2018.

\bibitem{szegedy2017inception}
C. Szegedy, S. Ioffe, V. Vanhoucke, and A. Alemi.
\newblock {Inception-v4, Inception-ResNet and the Impact of Residual
  Connections on Learning}.
\newblock In {\em Proc. AAAI}, 2017.

\bibitem{teichmann2019d2r}
M. Teichmann, A. Araujo, M. Zhu, and J. Sim.
\newblock {Detect-to-Retrieve: Efficient Regional Aggregation for Image
  Search}.
\newblock In {\em Proc. CVPR}, 2019.

\bibitem{tolias2015image}
G. Tolias, Y. Avrithis, and H. Jegou.
\newblock {Image Search with Selective Match Kernels: Aggregation Across Single
  and Multiple Images}.
\newblock {\em IJCV}, 2015.

\bibitem{tolias2020learning}
G. Tolias, T. Jenicek, and O. Chum.
\newblock {Learning and Aggregating Deep Local Descriptors for Instance-Level
  Recognition}.
\newblock In {\em Proc. ECCV}, 2020.

\bibitem{torii201524}
A. Torii, R. Arandjelović, J. Sivic, M. Okutomi, and T. Pajdla.
\newblock {24/7 place recognition by view synthesis}.
\newblock In {\em Proc. CVPR}, 2015.

\bibitem{WahCUB_200_2011}
C. Wah, S. Branson, P. Welinder, P. Perona, and S. Belongie.
\newblock {The Caltech-UCSD Birds-200-2011 Dataset}.
\newblock Technical Report CNS-TR-2011-001, California Institute of Technology,
  2011.

\bibitem{wang2018cosface}
H. Wang, Y. Wang, Z. Zhou, X. Ji, D. Gong, J. Zhou, Z. Li, and W. Liu.
\newblock {Cosface: Large Margin Cosine Loss for Deep Face Recognition}.
\newblock In {\em Proc. CVPR}, 2018.

\bibitem{wei2019rpc}
X. Wei, Q. Cui, L. Yang, P. Wang, and L. Liu.
\newblock {RPC: A Large-Scale Retail Product Checkout Dataset}.
\newblock {\em arXiv:1901.07249}, 2019.

\bibitem{Weinberger2006TripletLoss}
K.~Q. Weinberger, J. Blitzer, and L.~K. Saul.
\newblock {Distance metric learning for large margin nearest neighbor
  classification}.
\newblock In {\em Proc. NIPS}, 2006.

\bibitem{Weyand2015}
{Weyand, T. and Leibe, B.}
\newblock {Visual landmark recognition from Internet photo collections: A
  large-scale evaluation}.
\newblock {\em {Computer Vision and Image Understanding}}, 2015.

\bibitem{xie2017aggregated}
S. Xie, R. Girshick, P. Dollár, Z. Tu, and K. He.
\newblock {Aggregated Residual Transformations for Deep Neural Networks}.
\newblock In {\em Proc. CVPR}, 2017.

\bibitem{yan2017exploiting}
K. Yan, Y. Tian, Y. Wang, W. Zeng, and T. Huang.
\newblock {Exploiting Multi-Grain Ranking Constraints for Precisely Searching
  Visually-similar Vehicles}.
\newblock In {\em Proc. ICCV}, 2017.

\bibitem{Yokoo20Landmark}
S. Yokoo, K. Ozaki, E. Simo-Serra, and S. Iizuka.
\newblock {Two-stage Discriminative Re-ranking for Large-scale Landmark
  Retrieval}.
\newblock In {\em Proc. CVPR Workshops}, 2020.

\bibitem{zapletal2016vehicle}
D. Zapletal and A. Herout.
\newblock {Vehicle Re-Identification for Automatic Video Traffic Surveillance}.
\newblock In {\em Proc. CVPR}, 2016.

\bibitem{zhou2017places}
B. Zhou, A. Lapedriza, A. Khosla, A. Oliva, and A. Torralba.
\newblock {Places: A 10 million Image Database for Scene Recognition}.
\newblock {\em IEEE Transactions on Pattern Analysis and Machine Intelligence},
  2017.

\end{thebibliography}
}
}

\end{document}